\documentclass[lettersize,journal]{IEEEtran}

\usepackage{makecell}
\usepackage{multirow}
\usepackage[colorlinks]{hyperref}

\makeatletter
\newif\if@restonecol
\makeatother

\usepackage[ruled,vlined]{algorithm2e}
\usepackage{algpseudocode}
\usepackage{amsmath}
\renewcommand{\algorithmicrequire}{\textbf{Input: }}  

\usepackage{amsmath}
\usepackage{epsfig}
\usepackage{graphicx}
\usepackage{amsmath,amssymb} 
\usepackage{color}
\usepackage[switch,mathlines]{lineno}
\usepackage{array}
\usepackage{bbm}
\usepackage{subcaption}
\usepackage{mathrsfs}
\usepackage{verbatim}
\makeatletter
\DeclareRobustCommand\onedot{\futurelet\@let@token\@onedot}
\def\@onedot{\ifx\@let@token.\else.\null\fi\xspace}

\def\ie{\emph{i.e}\onedot,~}

\def\etal{\emph{et al}\onedot}
\def\sota{state-of-the-art~}
\def\aka{a.k.a\onedot~}
\newcommand{\printfnsymbol}[1]{%
  \textsuperscript{\@fnsymbol{#1}}%
}
\makeatother
\begin{document}

\title{Multi-Agent Semi-Siamese Training for \\ Long-tail and Shallow Face Learning}

\author{Hailin~Shi\printfnsymbol{1}\thanks{{*} Equal contribution.},~\IEEEmembership{Member,~IEEE,}
        Dan~Zeng\printfnsymbol{1},~\IEEEmembership{Member,~IEEE,}
        Yichun~Tai,
        Hang~Du,
        Yibo~Hu,
        Zicheng~Zhang,
        and~Tao~Mei,~\IEEEmembership{Fellow,~IEEE}
\thanks{Hailin Shi, Yibo Hu, and Tao Mei are with JD AI Research, Beijing 100020, China.
E-mail: shihailin@jd.com, tmei@live.com, huyibo871079699@gmail.com}
\thanks{Dan Zeng, Yichun Tai, and Hang Du are with Shanghai University, Shanghai 200444, China.
E-mail: \{dzeng, taiyc, duhang\}@shu.edu.cn}
\thanks{Zicheng Zhang is with University of Chinese Academy of Sciences, Beijing 100043, China.
E-mail: zhangzicheng19@mails.ucas.ac.cn}
\thanks{Corresponding author: Dan Zeng.}}



\maketitle

\begin{abstract}
With the recent development of deep convolutional neural networks and large-scale datasets, deep face recognition has made remarkable progress and been widely used in various applications.
However, unlike the existing public face datasets, in many real-world scenarios of face recognition, the depth of training dataset is shallow, which means only two face images are available for each ID.
With the non-uniform increase of samples, such issue is converted to a more general case, \textit{a.k.a} long-tail face learning, which suffers from data imbalance and intra-class diversity dearth simultaneously.
These adverse conditions damage the training and result in the decline of model performance.
Based on the Semi-Siamese Training (SST), we introduce an advanced solution, named Multi-Agent Semi-Siamese Training (MASST), to address these problems.
MASST includes a probe network and multiple gallery agents, the former aims to encode the probe features, and the latter constitutes a stack of networks that encode the prototypes (gallery features).
For each training iteration, the gallery network, which is sequentially rotated from the stack, and the probe network form a pair of semi-siamese networks.
We give the theoretical and empirical analysis that, given the long-tail (or shallow) data and training loss, MASST smooths the loss landscape and satisfies the Lipschitz continuity with the help of multiple agents and the updating gallery queue.
The proposed method is out of extra-dependency, thus can be easily integrated with the existing loss functions and network architectures.
It is worth noting that, although multiple gallery agents are employed for training, only the probe network is needed for inference, without increasing the inference cost.
Extensive experiments and comparisons demonstrate the advantages of MASST for long-tail and shallow face learning.
\end{abstract}

\begin{IEEEkeywords}
Face Recognition, Shallow Face Learning, Long-tail Face Learning.
\end{IEEEkeywords}

\section{Introduction}
\begin{figure}[ht]
    \centering
    \includegraphics[scale=0.3]{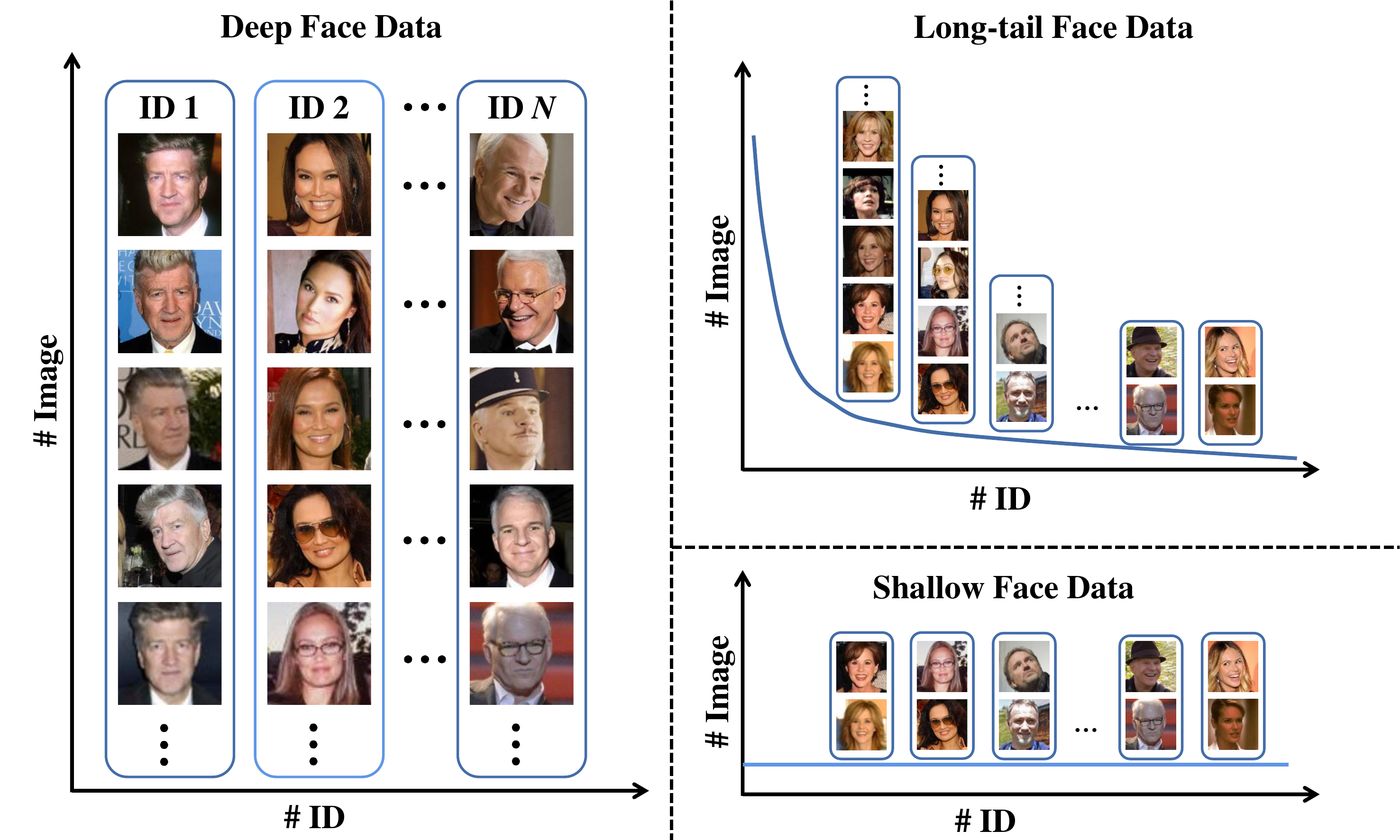}
    \caption{
    Left: in deep face data, each ID has plentiful samples, providing rich intra-class diversity.
    Top-right: long-tail face data.
    Bottom-right: shallow face data, which can be regarded as a special case of long-tail distribution, has a small number of samples per ID, and therefore the scarcity of intra-class diversity broadly exists in each ID.
    }
    \label{data}
\end{figure}

\IEEEPARstart{T}{he} convolutional neural networks (CNNs)~\cite{yoo2015attentionnet, schroff2015facenet, he2016deep} have shown great advantages in face recognition. Besides the advance of network design and training algorithm, the success largely relies on the large-scale training datasets, such as CASIA-WebFace~\cite{yi2014learning}, MS-Celeb-1M~\cite{guo2016ms}, VGGFace2~\cite{cao2018vggface2}, \textit{etc}. 
We term this type of dataset as deep face data, which provides abundant information both in breadth (IDs) and depth (samples per ID).
In many real-world scenarios, however, the training usually encounters unfavorable data distribution, such as shallow distribution~\cite{du2020semi}. 
The shallow face data distribution, which often has a large number of IDs, yet contains a few (two mostly) face images per ID (a registration photo and a spot photo, so-called ``gallery'' and ``probe''). 
According to the analysis in SST~\cite{du2020semi}, the lack of intra-class diversity prevents the network from effective optimization, and leads to the collapse of feature dimension.
Whereas, with the non-uniform increase of samples for each ID, the data tends to be long-tail distributed.
For example, in the large-scale face dataset MS-Celeb-1M, only a small portion of identities, about 10\%, have abundant face images, while a considerable portion has only a few samples.
The head classes contain enough samples with rich intra-class diversity to avoid the obvious imbalance problem in the learning process.
However, for tail classes, the amount of samples for each ID is too small to push the decision boundary correctly and classify with no bias, making the trained model prone to overfitting. 
As exemplified in Fig.~\ref{data}, shallow face data is an extreme case of long-tail face data, where it only has samples like tail data and the number of samples for each ID is two.
In such a situation, we find the existing training methods may not perform well.

With the further maturity of face recognition technology, more and more people begin to pay attention to the problems in real-world scenarios mentioned above.
The main challenging issues are as follows:
(1) Shallow face learning is similar to the existing problem of few-shot learning in face recognition~\cite{fei2006one}, which has been widely studied.
For example, to increase the sample size, some works are done to synthesize data based on generation rules~\cite{hariharan2017low,yin2018feature}. 
Some other works try to adjust the weight norm of the classifier for the few-shot classification~\cite{guo2017one}.
But there are still some differences between these two issues.
Most obviously, few-shot learning performs close recognition~\cite{guo2017one,wu2017low,cheng2017know,wang2018feature}, while shallow face learning focuses on the open-set recognition task in which test IDs are excluded from training IDs.
Besides, few-shot learning needs pre-training in the source domain (with deep data) before fine-tuning to the target domain~\cite{cheng2017know,zhu2019large,yin2019feature}, while shallow face learning tends to train on shallow face data from the very beginning.
Unfortunately, targeted and effective solutions have not been proposed.
(2) For face recognition tasks, some metric-based losses such as~\cite{schroff2015facenet,wen2016discriminative} are proved to be superior to conventional softmax loss.
On that basis, the range loss~\cite{zhang2017range}, which can enhance the discriminative power of the deeply learned features when dealing with few-shot samples in tail classes, is proposed to solve the long-tail problem.
As the inherent problems of long-tail distribution, the data imbalance and few-shot problem are concerned.
Most works related to the scarcity of samples rely on a base set with adequate samples and pay more attention to few-shot classification rather than learning a general representation under data with the long-tail distribution.
In addition, some classical strategies~\cite{khan2017cost,lin2017focal} against imbalance problem have been made full use in current deep learning frameworks, including resampling~\cite{chawla2002smote,he2008provably} and cost-sensitive learning~\cite{drummond2003c4,ting2000comparative}.
However, there are still many limitations such as discarding samples for under-sampling or introducing additional noises (\textit{e.g.}, assigning larger penalty on outlier samples).

To tackle the above challenges, we propose Multi-Agent Semi-Siamese Training (MASST) mechanism.
In long-tail face data, the tail identities cannot provide an accurate description with the limited number of training samples, and the feature space of them are often squeezed by the head identities.
Therefore, we randomly choose two photos in each ID, of which one photo is employed as the initial prototype, and the other one is employed as the training sample.
However, these two photos belong to the same ID and contain limited intra-class information.
To enlarge intra-class diversity, we design MASST to enforce the backbone being Semi-Siamese, which means the networks have close (but not identical) parameters.
As a result, MASST is proved to achieve better performance in shallow and long-tail face learning.

MASST consists of a probe network and multiple gallery agents.
For each ID in the training, the former extracts the feature from the probe as the training sample, and the latter sequentially rotated from the stack extracts the feature from the gallery as the prototype.
To guarantee the intra-class diversity between the feature, we combine moving-average with an auxiliary term to update the gallery network, which can constrain the slight difference among networks.
The gallery queue is built based on the gallery samples and updated in the training to replace the weights of the fully connected (FC) layer, which can reduce a large number of FC parameters.
As shown in Section \ref{EXPERIMENTS}, MASST achieves significant improvement in shallow and long-tail face learning.
It also outperforms the conventional training on the challenge of domain transfer.
Furthermore, since the proposed method is developed without extra-dependency, it can be flexibly integrated with the existing loss functions and network architectures.

The contributions of this paper are summarized as follows:
\begin{itemize}
\item Compared with shallow face learning, long-tail face learning is a more general case including the few-shot problem and the data imbalance problem at the same time. We study these two main challenges, which exist in many real-world scenarios of face recognition but have not been well addressed.
\item We propose Multi-Agent Semi-Siamese Training (MASST) method to address the few-shot problem and the data imbalance problem in shallow and long-tail face learning. MASST is able to achieve a smoother optimization process on the shallow and long-tail data by satisfying the Lipschitz continuity, and obtain the leading performance when it combines with the SOTA loss functions and network architectures flexibly.
\item We conduct comprehensive experiments to verify that MASST not only brings significant improvement on shallow and long-tail face learning, but also takes effect on both conventional deep data and domain transfer.
\end{itemize}

This paper is an extension of our previous conference version \cite{du2020semi}, and there are three major improvements over the preliminary one: 
(1) \textbf{MASST can tackle more general problems in face recognition.}
For the preliminary version, the Semi-Siamese Training (SST) is targeted proposed for shallow face learning.
As an extension of shallow distribution, the tail samples in long-tail distribution are shallow and the number of face images with different IDs varies greatly.
Thus, shallow face learning can be seen as a subproblem of long-tail face learning.
In the current version, we aim to address this more complicated issue, which is common in real-world scenarios of face recognition.
(2) \textbf{MASST has a smoother optimization process.}
The preliminary version designs SST, including a pair of Semi-Siamese networks, which have a probe network to embed the probe features, and a gallery network to update prototypes by gallery features.
On this basis, benefiting from the aforementioned Semi-Siamese structure, the gallery network of the current version is sequentially rotated from the network stack which involves multiple agents rather than the fixed single gallery network in SST.
At the same time, the update method of MASST is changed to achieve a smoother optimization process.
(3) \textbf{Theoretical analysis and comprehensive experiments are added.}
Through theoretical analysis, we prove that the networks in MASST can better satisfy the Lipschitz continuity.
Besides, comprehensive experiments on deep face data, shallow face data, and long-tail face data show the effectiveness of MASST. 
It can combine with the SOTA loss functions and network architectures flexibly and obtain the leading performance.

\section{Related Work}
\label{RELATED WORK}
For shallow and long-tail face learning, the main difficulties are the few-shot problem and the data imbalance problem.
So we review methods that solve these two issues in this section.
\subsection{Few-shot Learning}
The straightforward solution to enlarge the diversity of samples is data generation.
Hariharan and Girshick~\cite{hariharan2017low} propose to hallucinate additional samples of few-shot classes by category-independent transformations.
Dixitet \textit{et al.}~\cite{dixit2017aga} present an approach guided by attributes for data augmentation.
Delta-encoder~\cite{schwartz2018delta} is designed to synthesize new samples for unseen categories even if seeing few samples from them, by extracting transferable intra-class deformations and applying them to the few provided examples of a novel class.
Yin \textit{et al.}~\cite{yin2018feature} assume a Gaussian prior of the variance to transfer variations of normal classes to few-shot classes, which can augment the feature space of few-shot classes.
Ding \textit{et al.}~\cite{ding2019generative} propose a generative adversarial one-shot face recognizer to synthesize data for one-shot classes.
Specifically, a knowledge transfer generator and a general classifier are combined to guide the generation of effective data, which contributes to promoting the underrepresented classes.

Besides, to regularize the learning process, some methods adjust the conventional settings such as weight of samples and evaluation of loss.
Guo \textit{et al.}~\cite{guo2017one} develop the classifier to recognize few-shot samples by aligning the norms of the weight vectors of few-shot classes and the normal classes.
For few-shot learning in face recognition, Ma \textit{et al.}~\cite{ma2018hierarchical} adopt a hierarchical regularization method which utilizes coarse class labels for training and fine class labels for refining to avoid overfitting, Cheng \textit{et al.}~\cite{cheng2017know} propose the enforced softmax to guide model to produce a more compact vectorized representation by optimal dropout, selective attenuation, $L_2$ normalization, and model-level optimization.
Simon \textit{et al.}~\cite{simon2020adaptive} introduce a dynamic classifier that is constructed from few samples to design a framework for few-shot learning.
With such modeling, the results on the task of supervised and semi-supervised few-shot classification are competitive.

\subsection{Data Imbalance}
Arising from the long-tail distribution of nature data, the imbalance problem has been extensively studied~\cite{salakhutdinov2011learning,zhu2014capturing,liu2015deep,ouyang2016factors,cui2018large}.
Some metric-based methods add desirable constraints on the distance between samples in the feature space.
Huang \textit{et al.}~\cite{huang2016learning} propose quintuplet sampling with triple-header hinge loss to maintain both inter-cluster and inter-class margins for solving imbalanced distribution in vision classification tasks.
Center invariant loss~\cite{wu2017deep} aligns centers of the minority classes to the majority.
Range loss~\cite{zhang2017range} is developed to overcome the impact of long-tail distribution by simultaneously minimizing the harmonic mean of $k$ largest intra-class distance and maximizing the shortest inter-class distance within the mini-batch.
Ring loss~\cite{zheng2018ring} applies soft feature normalization to augment standard loss functions.
In addition, fair loss~\cite{liu2019fair} is proposed to balance different classes by using reinforcement learning.

In terms of data, some other methods have been proposed for improvement.
Classical strategies include under-sampling head classes, over-sampling tail classes, and data instance re-weighting, which aim to balance the number of samples.
By SMOTE~\cite{chawla2002smote}, over-sampling of the few-shot class and under-sampling of the normal class can be combined to achieve better classifier performance than replication-based methods.
BalanceCascade~\cite{liu2008exploratory} trains the learners sequentially.
The majority of class examples classified correctly are not taken into consideration further.
Unfortunately, these methods can easily suffer from over-fitting and introduce undesirable noises.
The under-sampling also might discard a large portion of samples so that missing valuable information.
Some of these solutions have been further extended to solve the data imbalance problem in deep learning~\cite{buda2018systematic}.
In order to remove representation bias, REPAIR~\cite{li2019repair} learns weights for different classes to re-sample data. 
Zhong \textit{et al.}~\cite{zhong2019unequal} attempt to treat the head data and the tail data of long-tail distribution in an unequal way to make full use of their respective characteristics.
Accompanying with noise-robust loss functions, these two training streams offer complementary information to deep feature learning.
Zhang \textit{et al.}~\cite{zhang2020class} focus on the optimization of dataset structures and build a medium-scale face recognition training set BUPT-CBFace, which can alleviate the problem of recognition bias.
Huang \textit{et al.}~\cite{huang2019deep} conduct extensive and systematic experiments to validate the effectiveness of classic schemes mentioned above.
Furthermore, CLMLE is proposed to reduce the class imbalance inherent in the data neighborhood, by enforcing a deep network to maintain inter-cluster margins both within and between classes.
As an extension of~\cite{ma2018hierarchical}, the two-step training schema~\cite{ma2020learning} is designed, which aggregates similar classes in tail parts and then selectively disperses those aggregated superclasses, to address the long-tail problem.

\section{Proposed Method}
\label{sec: PROPOSED METHOD}

In this section, we begin with the discussion on the major problems in shallow and long-tail face data.
Then, we introduce the method of Semi-Siamese and Multi-Agent Semi-Siamese Training.

\subsection{SST for Shallow Data}
\label{SST for shallow data}

The major issue in shallow face data is the lack of intra-class information.
Its adverse effect on model training has been discussed in Du~\etal~\cite{du2020semi}.
Here, we elaborate the detailed behavior of model when trained on shallow data with the current \sota methods, and introduce a novel perspective for understanding the difficulty of training on shallow face data and, more general cases, long-tail face data.

The \sota training methods are mostly developed from the softmax loss and its variants. 
Without loss of generality, taking softmax as an example, the formulation can be written as 
\begin{equation}
\mathcal{L}= -\log \frac{e^{s\,\boldsymbol{w_y}^T \boldsymbol{x}}}{e^{s\,\boldsymbol{w_y}^T \boldsymbol{x}}+\sum_{j=1,j\neq y}^{n} e^{s\,\boldsymbol{w_j}^T \boldsymbol{x}}},
\label{cross-entropy loss}
\end{equation}
where ${n}$ is the class number, $s$ is the scaling parameter, $\boldsymbol{x}$ is the sample feature with the L2 normalization ($\boldsymbol{x}=\boldsymbol{x}/\Vert \boldsymbol{x}\Vert_2$), $\boldsymbol{w_j}$ is the $j$-th class weight with the L2 normalization ($\boldsymbol{w_j}=\boldsymbol{w_j}/\Vert \boldsymbol{w_j}\Vert_2$), $y$ is the ground-truth label of the sample, and the output of the last FC layer 
is the inner product of the sample feature $x$ and $j$-th class weight $\boldsymbol{w_j}$. 
In the training process, the goal is maximizing the intra-class pair similarity $\boldsymbol{{w}_{y}}^{\mathrm{T}}\boldsymbol{x}$ and minimizing the inter-class pairs $\boldsymbol{{w}_{j}}^{\mathrm{T}}\boldsymbol{x}$.
If the sample number per class is large enough (\aka deep data), $\boldsymbol{x}$'s can provide abundant intra-class diversity.
There are, however, only two samples per class (\ie the gallery $\boldsymbol{x_g}$ and probe $\boldsymbol{x_p}$) in shallow data.
As the prototype~$\boldsymbol{{{w}_{y}}}$ is determined by $\boldsymbol{x_g}$ and $\boldsymbol{x_p}$, the three vectors will easily converge close ($\boldsymbol{w_y} \approx \boldsymbol{x_g} \approx \boldsymbol{x_p}$) that the loss value in this class approaches to the lower bound.
Therefore, in each iteration of batch-wise training, the network is well-fitted on a small number of classes and badly-fitted on the remaining ones.
As a result, the total loss value will be oscillating and the training will be harmed.
Moreover, due to the lack of the intra-class diversity in features space of all the classes ($\boldsymbol{x_g} \approx \boldsymbol{x_p}$), the prototype $\boldsymbol{{{w}_{y}}}$ is pushed to zeros in most dimensions so that it is unable to span an effective feature space.
Specifically, given a network and loss function in conventional training, the loss landscape will become much more winding when the data depth becomes shallow.
One can refer to Du~\etal~\cite{du2020semi} for detailed discussion.

In this context, SST is designed to improve the stability of training on shallow data.
We denote gallery and probe images by $I_g$ and $I_p$, and their features $x_g = \phi (I_g)$ and $x_p = \phi (I_p)$, where $\phi$ is the backbone.
$\phi$ should be Semi-Siamese, coined gallery network $\phi_g$ and probe network $\phi_p$, to maintain the distance between $\boldsymbol{x_g}$ and $\boldsymbol{x_p}$.
$\phi_g$ and $\phi_p$ have the same architecture and close (but non-identical) parameters, $\phi_g = \phi_p + \boldsymbol{\epsilon'}$, to prevent features from collapse, $\phi_g(I_g) = \phi_p(I_p) + \boldsymbol{\epsilon}$.
To implement the Semi-Siamese networks, a common approach is to add a network constraint $\| \phi_g - \phi_p \| < \epsilon'$ in the training loss.
As suggested by MoCo~\cite{he2020momentum}, we choose a better approach that updating gallery network in the momentum way, which is defined as
\begin{equation}
{\phi}_{g} = {m}{\phi}_{g}+{(1-m)}{\phi}_{p},
\label{moving_average}
\end{equation}
where $m$ is the weight of moving-average, and the probe network $\phi_p$ updates with SGD. 

The prototype constraint is deployed in the training objective to enlarge the entries of prototype, such like $\mathcal{L} + \beta (\alpha - \|\boldsymbol{w_y}\|)$ with parameters $\alpha$ and $\beta$, avoiding the vanishing issue in dimensions of $\boldsymbol{w_y}$. 
This technique, however, does not show effective improvement (Table~\ref{tbl:ablation_study}), since the entries are enlarged uniformly.
To handle this issue, SST replaces $\boldsymbol{w_y}$ with the gallery feature $\boldsymbol{x_g}$ as the prototype rather than manipulating $\boldsymbol{w_y}$ itself.
The feature-based prototype avoids the vanishing issue, and thereby keeps the discriminative components in the prototype.


\subsection{From SST to MASST}
\label{From SST to MASST}

The experiments in Du~\etal~\cite{du2020semi} show that SST has remarkable advantage over the conventional training on shallow data.
Inspired by Markov-Lipschitz theory~\cite{li2020markov}, we analyze the advantage of SST from the perspective of the bi-Lipschitz continuity in the following.

A function $\eta:\mathbb{R}^{H \times W} \rightarrow\mathbb{R}^n$ 
is locally bi-Lipschitz if for all $x_j \in \mathcal{N}_i$ (the neighborhood of $x_i$), there exists a real constant $K\geq1$ satisfying 
\begin{equation}
{\Vert x_i-x_j \Vert}_2/K \leq{\Vert \eta(x_i)-\eta(x_j) \Vert}_2 \leq K{\Vert x_i-x_j \Vert}_2.
\label{bi-Lipschitz}
\end{equation}
$K$ is termed as bi-Lipschitz constant.
According to Convex Optimization theory~\cite{nesterov2018lectures}, a differentiable loss function $\mathcal{L}(x)$ is called smooth if its gradient $\eta(x)$ is Lipschitz or bi-Lipschitz continuous.
The optimization of smooth function will be more stable and robust if $K$ is smaller~\cite{li2020markov}.



First, we compare the bi-Lipschitz constant $K$ of the conventional training and SST to compare their stability of optimization process.
For conventional training, according to Eq.~\ref{cross-entropy loss}, 
when the gallery $I_g$ and probe $I_p$ are fed into the backbone $\phi$ separately, we have the gradient

\begin{equation}
\begin{split}
\eta_{Convt.}(I_g)
=s  (\frac{e^{s\,\boldsymbol{w_y}^T \phi(I_g)}}{\sum_{j=1}^{n} e^{s\,\boldsymbol{w_j}^T \phi(I_g)}}-1)\phi(I_g),\\
\eta_{Convt.}(I_p) 
=s (\frac{e^{s\,\boldsymbol{w_y}^T \phi(I_p)}}{\sum_{j=1}^{n} e^{s\,\boldsymbol{w_j}^T \phi(I_p)}}-1)\phi(I_p) .
\end{split}
\label{convt_Ig_Ip}
\end{equation}
As the training carries on, for $I_g$ and $I_p$ that belong to the same ID, $\phi(I_g) \approx \phi(I_p)$ and $\eta_{Convt.}(I_g) \approx \eta_{Convt.}(I_p)$.
This means $K$ is very large.
Thereby, the optimization of conventional training on shallow data is difficult to be smooth and steady~\cite{li2020markov}.                                         

As for SST, $\boldsymbol{w_y}$ is replaced with the gallery feature $\boldsymbol{x_g} = \phi_g(I_g)$, and the gradient with respect to $\boldsymbol{x_g}$ becomes to
\begin{equation}
\begin{aligned}
\eta_{x_g}(\boldsymbol{x})=s (\frac{e^{s\,\boldsymbol{x_g}^T \boldsymbol{x}}}{\sum_{j=1}^{n} e^{s\,\boldsymbol{x_j}^T \boldsymbol{x}}}-1)\boldsymbol{x}. 
\label{gradient of sst}
\end{aligned}
\end{equation}
Given the shallow data including a pair of gallery and probe per ID, which are the input of SST, Eq.~\ref{gradient of sst} is developed to
\begin{equation}
\begin{aligned}
\eta_{x_g}&([\boldsymbol{x_g};\boldsymbol{x_p}])\\
&= s(\frac{e^{s\,\boldsymbol{x_g}^T \boldsymbol{x_p} }}{e^{s\,\boldsymbol{x_g}^T \boldsymbol{x_p} }+\sum_{j=1,j\neq y}^{n} e^{s\,\boldsymbol{x_j}^T \boldsymbol{x_p}}}-1)\boldsymbol{x_p},\\
\end{aligned}
\end{equation}
where $\boldsymbol{x_g}=\phi_g(I_g)$ and $\boldsymbol{x_p}=\phi_p(I_p)$.


For the clarity of the following deduction, we denote $f$ as a function of $\boldsymbol{x_g}$ and $\boldsymbol{x_p}$ that
\begin{equation}
\begin{aligned}
f([\boldsymbol{x_g};\boldsymbol{x_p}]) = \frac{e^{s\,\boldsymbol{x_g}^T \boldsymbol{x_p}}}{e^{s\,\boldsymbol{x_g}^T \boldsymbol{x_p}}+\sum_{j=1,j\neq y}^{n} e^{s\,\boldsymbol{x_j}^T \boldsymbol{x_p}}}-1.
\end{aligned}
\end{equation}

Then, the upper bound of the variation of $\eta_{x_g}$ is computed as

\begin{equation}
\begin{aligned}
&\lVert \eta_{x_g}([\boldsymbol{x_g};\boldsymbol{x_p}])-\eta_{x_g}([\boldsymbol{x_g'};\boldsymbol{x_p}]) \rVert_2\\ 
&= \lVert s f([\boldsymbol{x_g};\boldsymbol{x_p}])\boldsymbol{x_p}-s f([\boldsymbol{x_g'};\boldsymbol{x_p}])\boldsymbol{x_p} \rVert_2\\
&=\lVert s(f([\boldsymbol{x_g};\boldsymbol{x_p}])-f([\boldsymbol{x_g'};\boldsymbol{x_p}])) \boldsymbol{x_p} \rVert_2\\
&=s \lVert f([\boldsymbol{x_g};\boldsymbol{x_p}])-f([\boldsymbol{x_g'};\boldsymbol{x_p}])\rVert_2\\
&=s\lVert \nabla_{\boldsymbol{x_g}} f^{T}_{\boldsymbol{x_g}; \boldsymbol{x_p}}(\boldsymbol{x_g'}-\boldsymbol{x_g})+o(\lVert \boldsymbol{x_g'}-\boldsymbol{x_g} \rVert_2)\rVert_2\\
&= s\lVert \boldsymbol{x_g}-\boldsymbol{x_g'}\rVert_2\lVert \nabla_{\boldsymbol{x_g}} f^{T}_{\boldsymbol{x_g}; \boldsymbol{x_p}}\rVert_2\\
&\leq \frac{s^2}{2}\lVert \boldsymbol{x_g}-\boldsymbol{x_g'}\rVert_2,
\end{aligned}
\end{equation}
where $\boldsymbol{x_g'} = \boldsymbol{x_g} +\epsilon_1$. Then, we can obtain the ratio to the input variation is

\begin{equation}
\begin{aligned}
\label{K of SST}
\frac{\lVert \eta_{x_g}([\boldsymbol{x_g};\boldsymbol{x_p}])-\eta_{x_g}([\boldsymbol{x_g'};\boldsymbol{x_p}]) \rVert_2}{\lVert \boldsymbol{x_g}-\boldsymbol{x_g'}\rVert_2}\leq \frac{s^2}{2}.
\end{aligned}
\end{equation}
One can refer to the supplementary material for the deduction details.
This indicates that there always exists a constant $K$ making SST satisfies the right side inequality in Eq.~\ref{bi-Lipschitz}.
Besides, the ratio of variation also has a lower bound. This is because that $\eta_{x_g}([\boldsymbol{x_g};\boldsymbol{x_p}]) \neq \eta_{x_g}([\boldsymbol{x'_g};\boldsymbol{x_p}])$ always holds owing to the Semi-Siamese structure and the momentum-updating method.

In most of the existing experiments, the parameter $s$ is set around 30, so $\frac{s^2}{2} > 1$.
Therefore, there exists $K\geq 1$ that constrains the variation range of ${\Vert \eta_{x_g}([\boldsymbol{x_g};\boldsymbol{x_p}])-\eta_{x_g}([\boldsymbol{x_g'};\boldsymbol{x_p}]) \Vert}_2$. In contrast to the conventional training, SST can satisfy the bi-Lipschitz continuity, which is the basic reason that SST achieves stable and robust convergence on shallow data.

Based on the advantage of SST on shallow data, we further extend it to handle a more general problem that is long-tail face learning. 
Unfortunately, with the non-uniform distribution of training samples in long-tail face data, SST suffers from bad convergence.
The tail classes have faster convergence speed due to the fewer samples, but the head classes including large number of samples converge slower.
The power of SST is undermined on long-tail data.

\begin{figure}[ht]
    \centering
    \includegraphics[scale=0.46]{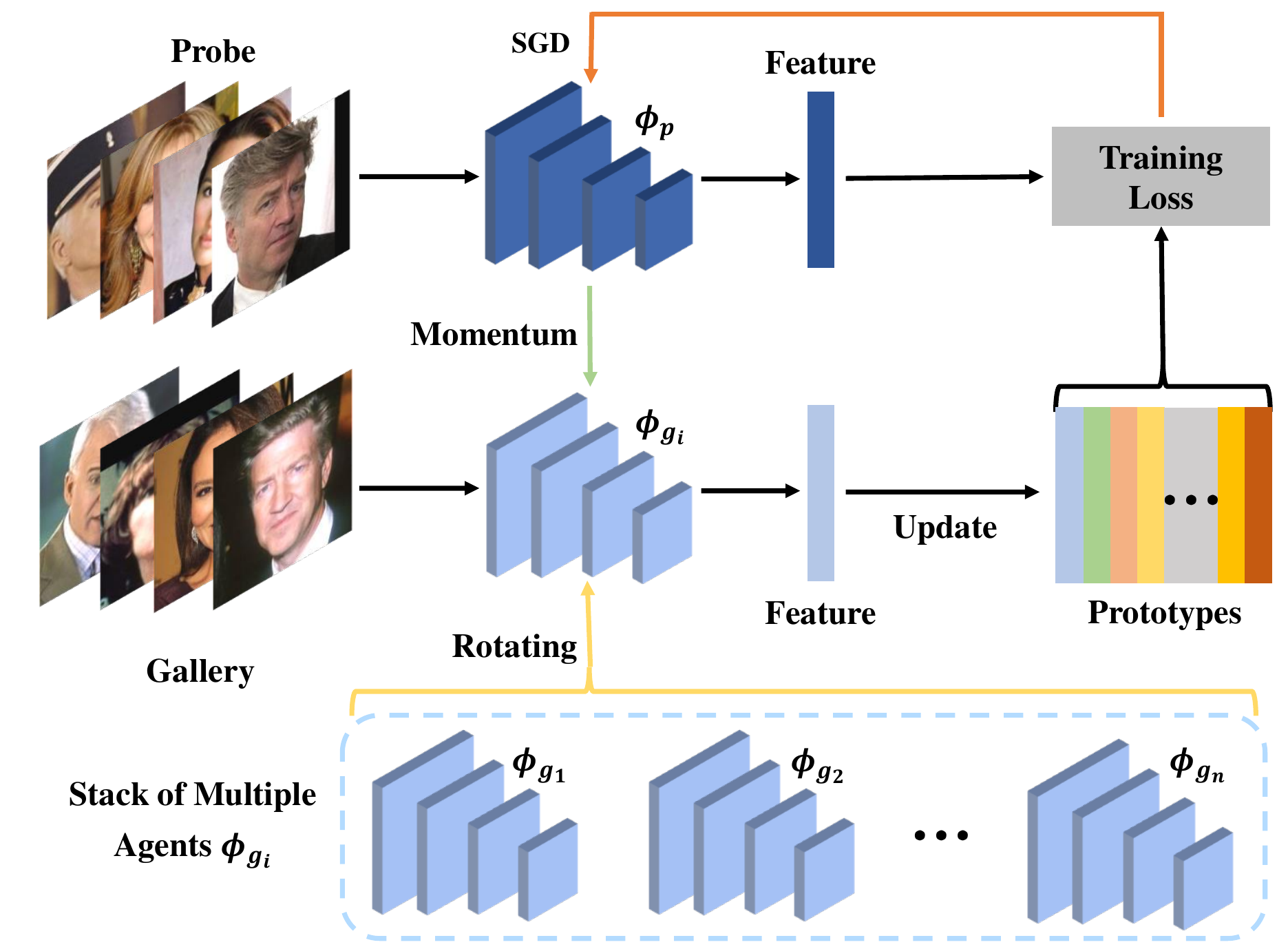}
    \caption{
    Illustration of Multi-Agent Semi-Siamese Training. It consists of a probe network ($\phi_p$) to encode the probe features, and multiple gallery agents (${\phi_g}_i$), each of which plays the role of gallery network alternatively, to update prototypes by gallery features. 
    The probe features and the feature-based prototypes are employed to compute the training loss. Then, the probe network is optimized via SGD w.r.t. the training loss (the orange arrow), and the gallery network (\textit{i.e} the current agent) is updated in the momentum way (the green arrow). Best viewed in color. }
    \label{fig:MASST}
\end{figure}

In order to better satisfy the locally bi-Lipschitz continuity, we propose MASST to strengthen the stability and robustness with smaller value of $K$.
The single gallery network in SST is replaced with a gallery network stack including multiple agents ($\phi_{g_1},\phi_{g_2},\cdots,\phi_{g_{S}}$).
The framework of MASST is depicted in Fig.~\ref{fig:MASST}.
In the training process, we sequentially rotate one agent from the stack as the gallery network.
The update method of the $i$-th agent is defined as
\begin{equation}
{\phi}_{g_i} = (1+a)(m{\phi}_{g_i}+(1-m){\phi}_{p})-a\frac{\sum_{j=1,j\neq i}^{S}{\phi}_{g_j}}{S-1},
\label{MASST_update}
\end{equation}
where $i=1,2,\cdots,S$, $S$ is the number of agents in the gallery network stack, and $a$ is a parameter that adjusts the weight of the first and second terms.
With the second term, the agents in the gallery network stack can maintain distance intentionally.
Therefore, MASST can have smaller value of $K$ to achieve smoother optimization process.


Similar to SST, we have the gradient in MASST 
\begin{equation}
\begin{aligned}
&\eta_{x_g}([\boldsymbol{{x_g}^{ma}};\boldsymbol{{x_p}^{ma}}])\\
&= s(\frac{e^{s\,\boldsymbol{{x_g}^{ma}}^T \boldsymbol{{x_p}^{ma}} }}{e^{s\,\boldsymbol{{x_g}^{ma}}^T \boldsymbol{{x_p}^{ma}} }+\sum_{j=1,j\neq y}^{n} e^{s\,\boldsymbol{{x_j}^{ma}}^T \boldsymbol{{x_p}^{ma}}}}-1)\boldsymbol{{x_p}^{ma}},\\
\end{aligned}
\end{equation}
where $\boldsymbol{{x_g}^{ma}}=\phi_{g_i}(I_g)$ and $\boldsymbol{{x_p}^{ma}}=\phi_p(I_p)$. 

Without loss of generality, ignoring the scaling parameter $s$, the ratio of the variation of $\eta_{x_g}([\boldsymbol{x_g};\boldsymbol{x_p}])$ to the variation of $\eta_{x_g}([\boldsymbol{{x_g}^{ma}};\boldsymbol{{x_p}^{ma}}])$ can be written as
\begin{equation}
\begin{aligned}
&\frac{\lVert \eta_{x_g}([\boldsymbol{{x_g}^{ma}};\boldsymbol{{x_p}^{ma}}])-\eta_{x_g}([\boldsymbol{{x_g'}^{ma}};\boldsymbol{{x_p}^{ma}}]) \rVert_2}{\lVert \eta_{x_g}([\boldsymbol{x_g};\boldsymbol{x_p}])-\eta_{x_g}([\boldsymbol{x_g'};\boldsymbol{x_p}]) \rVert_2}\\
&=\frac{\lVert \nabla_{\boldsymbol{{x_g}^{ma}}}\frac{e^{\boldsymbol{{x_g}^{ma}}^T \boldsymbol{{x_p}^{ma}}}}{e^{\boldsymbol{{x_g}^{ma}}^T \boldsymbol{{x_p}^{ma}}}+e^{\boldsymbol{{x_g'}^{ma}}^T \boldsymbol{{x_p}^{ma}}}}\rVert_2}{\lVert \nabla_{\boldsymbol{{x_g}}}\frac{e^{\boldsymbol{{x_g}}^T \boldsymbol{{x_p}}}}{e^{\boldsymbol{{x_g}}^T \boldsymbol{{x_p}}}+e^{\boldsymbol{{x_g'}}^T \boldsymbol{{x_p}}}}\rVert_2},
\end{aligned}
\end{equation}
where $\boldsymbol{{x_g'}^{ma}} = \boldsymbol{{x_g}^{ma}}  +\epsilon_2$, and $\lVert \nabla_{\boldsymbol{{x_g}}}\frac{e^{\boldsymbol{{x_g}}^T \boldsymbol{{x_p}}}}{e^{\boldsymbol{{x_g}}^T \boldsymbol{{x_p}}}+e^{\boldsymbol{{x_g'}}^T \boldsymbol{{x_p}}}}\rVert_2$ can be developed to 
\begin{equation}
\begin{aligned}
\label{monotone}
&\lVert \nabla_{\boldsymbol{{x_g}}}\frac{e^{\boldsymbol{{x_g}}^T \boldsymbol{{x_p}}}}{e^{\boldsymbol{{x_g}}^T \boldsymbol{{x_p}}}+e^{\boldsymbol{{x_g'}}^T \boldsymbol{{x_p}}}}\rVert_2\\
&=\lVert \frac{{e^{\boldsymbol{{x_g}}^T \boldsymbol{{x_p}}}e^{\boldsymbol{{x_g'}}^T \boldsymbol{{x_p}}}}}{({e^{\boldsymbol{{x_g}}^T \boldsymbol{{x_p}}}+e^{\boldsymbol{{x_g'}}^T \boldsymbol{{x_p}}}})^2}\boldsymbol{{x_p}}\rVert_2\\
&=\lVert\frac{1}{e^{\boldsymbol{{(x_g-x_g')}}^T\boldsymbol{{x_p}}}+e^{\boldsymbol{{(x_g'-x_g)}}^T\boldsymbol{{x_p}}}+2}\rVert_2.
\end{aligned}
\end{equation}

Compared with SST, the update method of the gallery network in MASST increases differences among agents in the stack, \textit{i.e.}, $\epsilon_2 > \epsilon_1$.
Then, we can find that
\begin{equation}
\begin{aligned}
\boldsymbol{{(x_g-x_g')}}^T&\boldsymbol{{x_p}}\\
&\leq \lVert\boldsymbol{{(x_g-x_g')}}\rVert_2\lVert \boldsymbol{{x_p}}\rVert_2=\epsilon_1,\\
\boldsymbol{{({x_g}^{ma}-{x_g'}^{ma})}}^T&\boldsymbol{{{x_p}^{ma}}}\\
&\leq \lVert\boldsymbol{{({x_g}^{ma}-{x_g'}^{ma})}}\rVert_2\lVert \boldsymbol{{{x_p}^{ma}}}\rVert_2=\epsilon_2.
\end{aligned}
\end{equation}
According to the regional monotonicity of Equation~\ref{monotone} \textit{w.r.t.} $\boldsymbol{{(x_g-x_g')}}^T\boldsymbol{{x_p}}$ (more details are shown in the supplementary material), we obtain:
\begin{equation}
\begin{aligned}
\frac{\lVert \eta_{x_g}([\boldsymbol{{x_g}^{ma}};\boldsymbol{{x_p}^{ma}}])-\eta_{x_g}([\boldsymbol{{x_g'}^{ma}};\boldsymbol{{x_p}^{ma}}]) \rVert_2}{\lVert \eta_{x_g}([\boldsymbol{x_g};\boldsymbol{x_p}])-\eta_{x_g}([\boldsymbol{x_g'};\boldsymbol{x_p}]) \rVert_2}\leq 1,
\end{aligned}
\end{equation}
for $\boldsymbol{{(x_g-x_g')}}^T\boldsymbol{{x_p}}\leq \boldsymbol{{({x_g}^{ma}-{x_g'}^{ma})}}^T\boldsymbol{{{x_p}^{ma}}}$.
By further comparing the ratio to the input variation in SST and MASST,
\begin{equation}
\begin{aligned}
&\frac{\lVert \eta_{x_g}([\boldsymbol{{x_g}^{ma}};\boldsymbol{{x_p}^{ma}}])-\eta_{x_g}([\boldsymbol{{x_g'}^{ma}};\boldsymbol{{x_p}^{ma}}]) \rVert_2\cdot\lVert \boldsymbol{x_g}-\boldsymbol{x_g'}\rVert_2}{\lVert \eta_{x_g}([\boldsymbol{x_g};\boldsymbol{x_p}])-\eta_{x_g}([\boldsymbol{x_g'};\boldsymbol{x_p}]) \rVert_2\cdot\lVert \boldsymbol{{x_g}^{ma}}-\boldsymbol{{x_g'}^{ma}}\rVert_2}\\
&\leq \epsilon_1/\epsilon_2 < 1,
\end{aligned}
\end{equation}
We can verify that there exists a smaller constant $K$ making MASST better satisfies the bi-Lipschitz continuity than SST.
In Section~\ref{MASST for long-tail data}, the result of experiment also proves this property of MASST.


It is worth noting that, although multiple gallery agents are involved for training, \textbf{we only utilize the probe network for test inference.}
Thereby, the test accuracy is improved while keeping the inference cost unchanged.

\subsection{MASST for Long-tail Data}
\label{MASST for long-tail data}
With the non-uniform increase of samples, the training suffers from data imbalance and intra-class diversity dearth simultaneously, which leads to the instability of the convergence process.

In order to verify the effectiveness of MASST in long-tail face learning, we construct long-tail data by the existing identity-based sampling method.
It is expressed as follows,
\begin{equation}
Num_{new} = Num_{org} \cdot (index + 1)^{-r},
\label{long-tail sampling}
\end{equation}
where $index$ is the ID index sorted in descending order according to the number of intra-class samples that $index \in N^*$, $Num_{org}$ and $Num_{new}$ are the numbers of intra-class samples before and after sampling, 
and $r$ is an adjusting parameter.

\begin{figure}[ht]
\centering
\begin{center}
\begin{minipage}[ht]{0.48\linewidth}
\includegraphics[width=4.2cm]{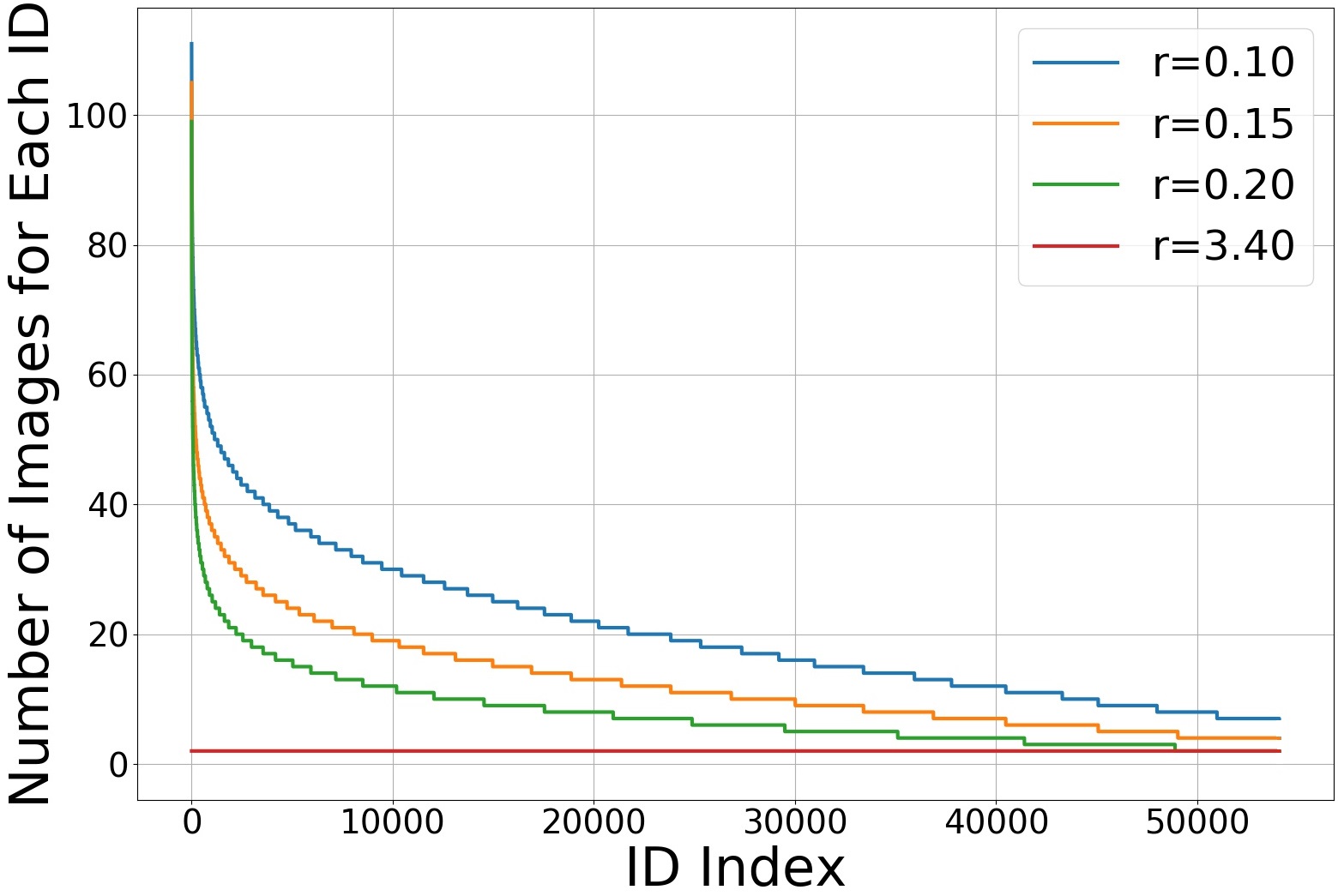}
\subcaption{}
\label{num/id}
\end{minipage}
\begin{minipage}[ht]{0.5\linewidth}
\includegraphics[width=4.4cm]{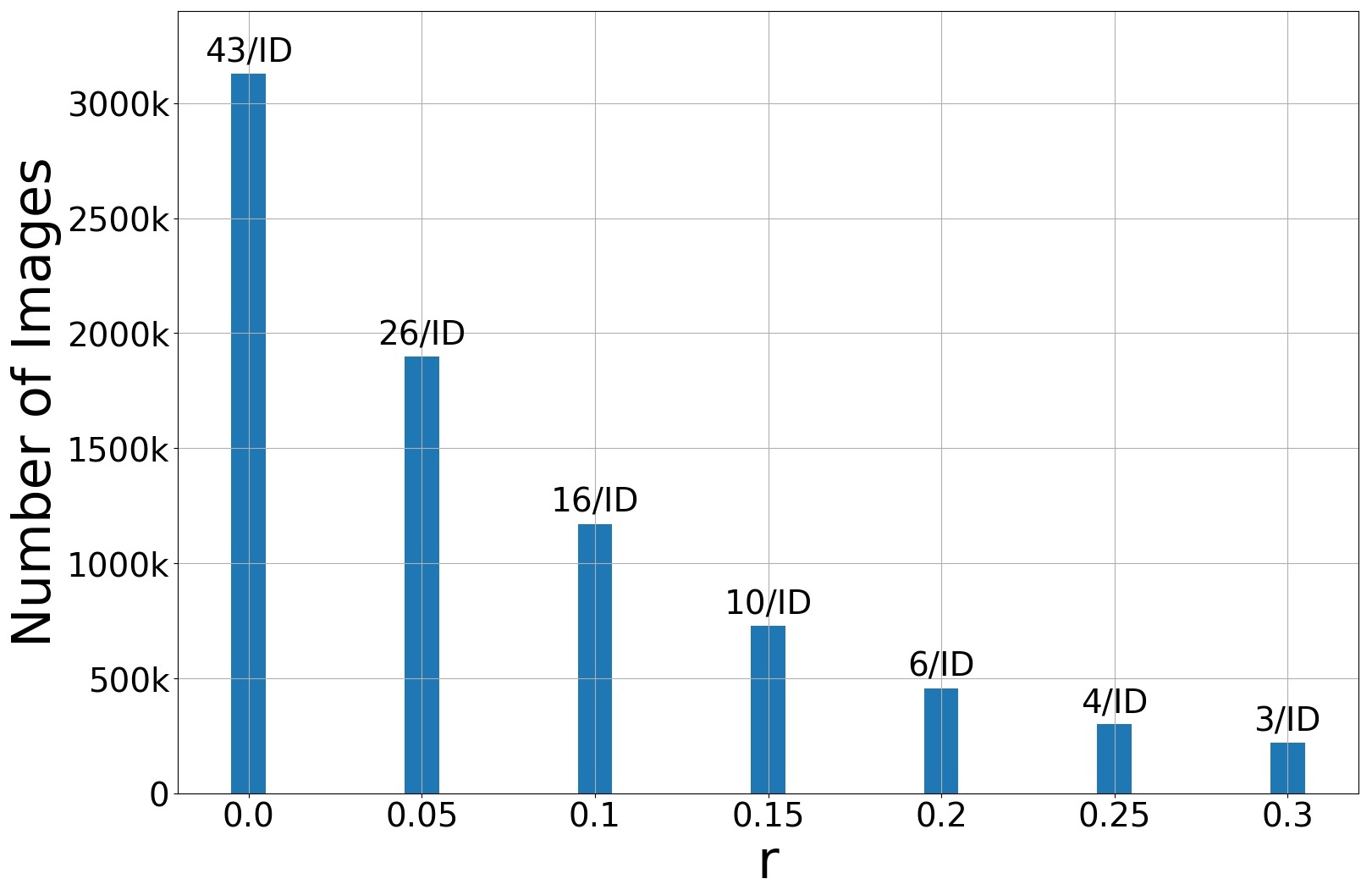}
\subcaption{}
\label{num}
\end{minipage}
\end{center}

\caption{The distribution of synthetic long-tail face data with different $r$.
(a) With the increase of $r$, the degree of long-tail distribution is intensified, and the depth of data becomes shallow finally. (b) The total number of images decreases as $r$ increases. The digits above the bar are the averaged number of images per ID.}
\label{long-tail data}
\end{figure}

With the increase of $r$, the degree of long-tail distribution is intensified and the sample size becomes smaller correspondingly.
$Num_{new} \ge 2$ guarantees that there are at least two images as input for training in tail classes.
Different from deep data, the distribution of synthetic long-tail face data is non-uniform (as shown in Fig.~\ref{long-tail data} (a)). 
Obviously, shallow face data is an extreme case of long-tail face data, and long-tail face learning is a more general case, which suffers from data imbalance and the lack of intra-class diversity simultaneously.

To compare the optimization process, the key problem here is measuring the loss landscape's curvature.
By means of the Taylor expansion, we have
\begin{equation}
\eta(x_i)=\eta(x_j)+\frac{\partial\eta(x_j)}{\partial x_j}(x_i -x_j)+o(\lVert x_i -x _j \rVert),
\end{equation}
then,
\begin{equation}
\begin{aligned}
{\Vert \eta(x_i)-\eta(x_j) \Vert}_2&\approx (x_i -x_j)^TH^TH(x_i -x_j)\\
&\leq \lVert H \rVert_2 \lVert x_i - x_j \rVert_2,
\end{aligned}
\end{equation}
where $H$ is the Hessian matrix of loss function $\mathcal{L}(x)$.


Therefore, the Lipschitz norm is given by ${\Vert \eta \Vert}_{Lip}$ = sup$\sigma(\nabla \eta)$ = sup$\sigma(H)$ = $\sigma(H)$, where $\sigma(A)$ is the spectral norm of the matrix $A$ ($L_2$ matrix norm of $A$).
The gradient $\eta$ satisfies the Lipschitz continuity better with the smaller spectral norm $\sigma(H)$.
Naively forming the Hessian matrix to compute the spectral norm has a prohibitive computational cost.
If there are $q$ parameters (weights and biases) in the network, then the Hessian matrix has dimensions $q \times q$.
The computational effort needed to evaluate the Hessian will scale like $O(q^2)$, and it also requires storage that is $O(q^2)$.
In order to compute the principal eigenvalue of the Hessian matrix efficiently, we use the power method, which is a matrix-free algorithm~\cite{yao2018large}.
For a random vector $v$, whose dimension is the same as $\eta$, as it is independent to $\theta$, we have
\begin{equation}
\frac{\partial(\eta^T v)}{\partial \theta} = \frac{\partial \eta^T}{\partial \theta} v + \eta^T \frac{\partial v}{\partial \theta} = \frac{\partial \eta^T}{\partial \theta} v = H v,
\label{Power Iteration for Eigenvalue Computation}
\end{equation}
where $H$ is the Hessian matrix.
The curvature can be computed efficiently in this way.

\begin{algorithm}
\caption{Power Iteration for Eigenvalue Computation}
\label{power_method}
\algorithmicrequire{Model Parameter $\theta$}.\\
Compute the gradient of $\mathcal{L}$ \textit{w.r.t.} $\theta$, \textit{i.e.}, $\eta=\frac{\mathrm{d} \mathcal{L} }{\mathrm{d} \theta}$.\\
Initialize a random vector $v$ (same dimension as $\eta$).\\
Normalize $v$, $v= \frac{v}{\lVert v \rVert_2}$.\\
\For {$i = 1,2,...,n$} 
{
Compute $\eta^T v$\\
Compute $Hv$ by backpropagation, $Hv = \frac{\mathrm{d} (\eta^Tv) }{\mathrm{d} \theta}$\\
Normalize and reset $v$, $v = \frac{Hv}{\lVert Hv \rVert_2}$
}
\end{algorithm}

The algorithm is shown in Alg.~\ref{power_method}.
Specifically, we first compute the gradient $\eta$, and initialize a random vector $v$, followed by the operation of the inner product.
Thus, we simply backpropagate $\eta^Tv$ rather than computing the full Hessian matrix.
When the error between two calculations of principal eigenvalue is less than the threshold value ($thr = 1e-3$), or the iterations exceed the maximum iterations ($max = 50$), the loop is stopped.
In this way, the quantity of interest is not the Hessian matrix $H$ itself but the product of $H$ with some vector $v$. 
The Hessian vector product $Hv$ that we wish to calculate, however, has only $q$ elements, so instead of computing the Hessian as an intermediate step, we find an efficient approach to evaluating $Hv$ directly that requires only $O(q)$ operations.

\begin{figure}[ht]
\centering
\begin{center}
\begin{minipage}[ht]{0.99\linewidth}
\includegraphics[width=8.5cm]{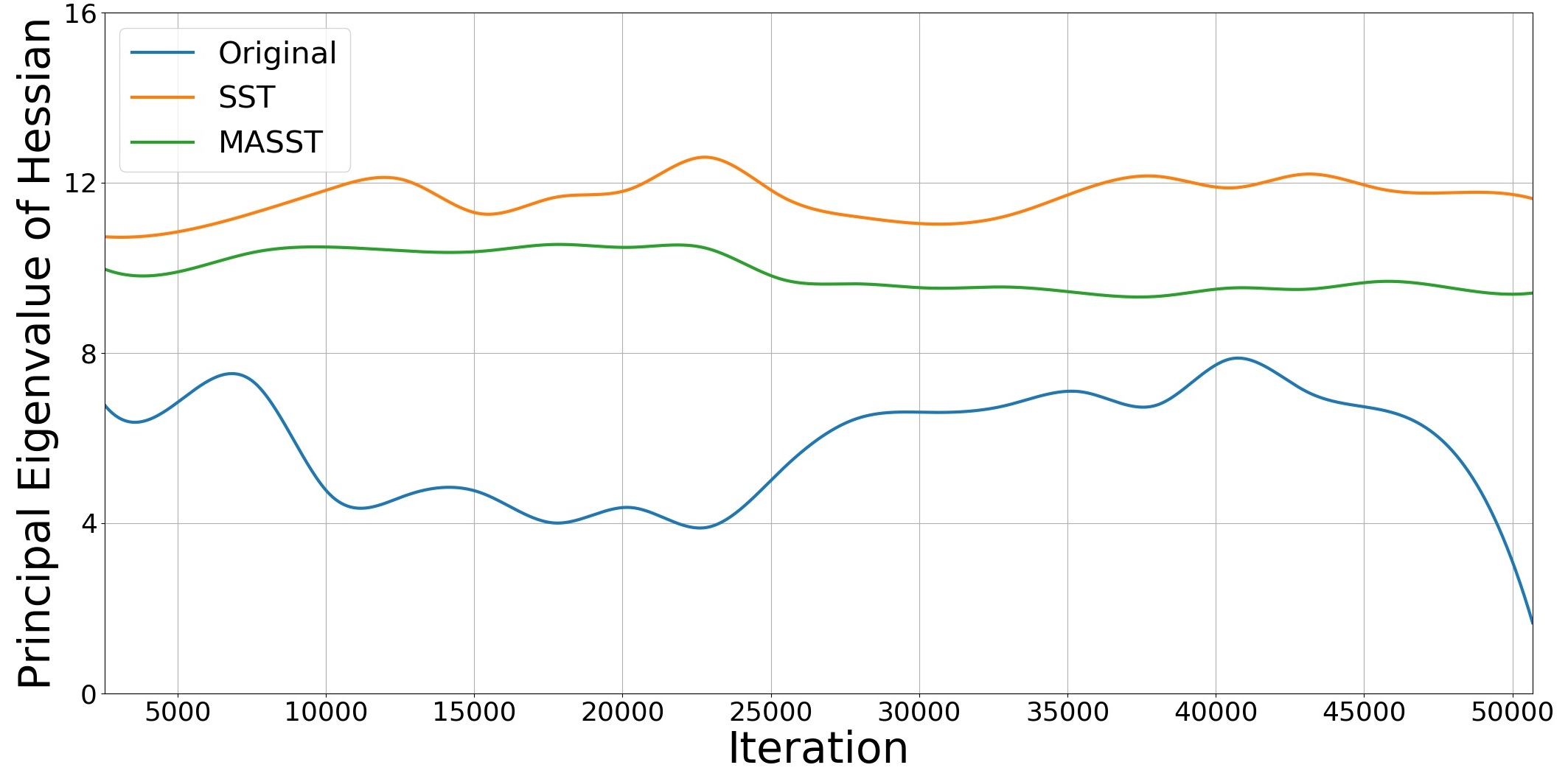}
\subcaption{}
\label{sn_shallow}
\end{minipage}
\begin{minipage}[ht]{0.99\linewidth}
\includegraphics[width=8.5cm]{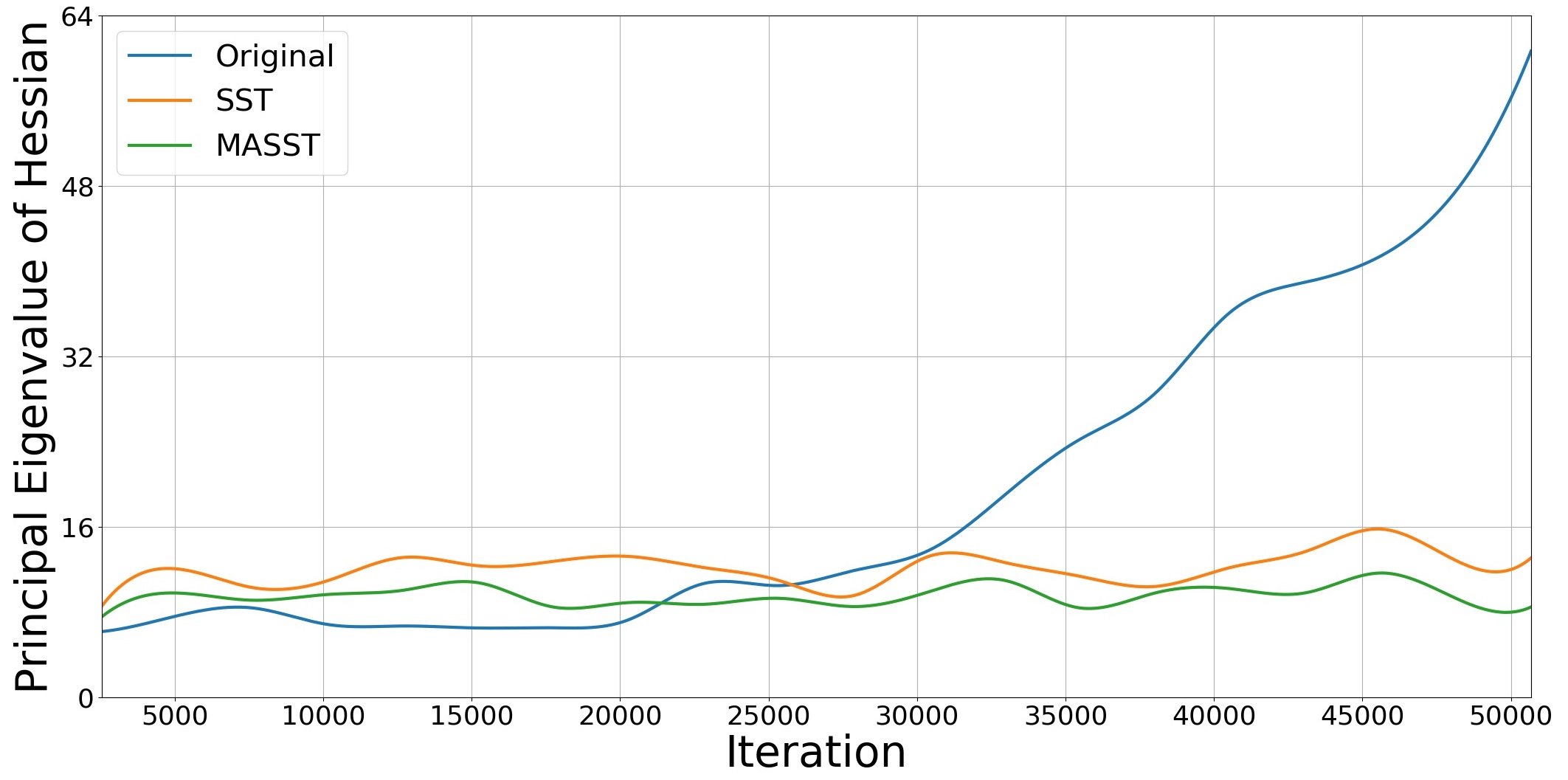}
\subcaption{}
\label{sn_longtail}
\end{minipage}
\end{center}

\caption{
(a) Comparison of the principal eigenvalue of the Hessian matrix on shallow face data. (b) Comparison of the principal eigenvalue of the Hessian matrix on long-tail face data. We can observe that the conventional training encounters either the over-small or over-large curvature on the training landscape, which indicates the problem of convergence towards the optimal solution.
Whereas, SST and MASST maintain the curvature at the medium range; moreover, MASST has more steady curvature than SST, which can be also verified by the standard deviation in Table~\ref{tb:sn}. Thereby, MASST empowers the model to accomplish better training.
}
\label{sn}
\end{figure}

For the conventional training, SST and MASST with MobileFaceNet on shallow face data and synthetic long-tail face data, we calculate the principal eigenvalue of the Hessian matrix during the training process.
As shown in Fig.~\ref{sn}, we can find: (1) the curvature of the conventional training becomes too small on shallow face data, which means it converges to the local minimum and optimizes invalidly; (2) on long-tail face data, the curvature of the conventional training gradually increases with the increase of iteration times, which means the training process becomes more tortuous; (3) compared with conventional training, the curvatures of SST and MASST are maintained at the medium range whether on shallow face data or long-tail face data.
We also calculate the average value and standard deviation of curvature during the training process in Table~\ref{tb:sn}.
On both shallow face data and long-tail face data, \textbf{MASST has a smaller average value and standard deviation of curvature to maintain the stable optimization process}.

\begin{figure}[ht]
\centering
\begin{center}
\begin{minipage}[ht]{0.48\linewidth}
\includegraphics[width=4.2cm]{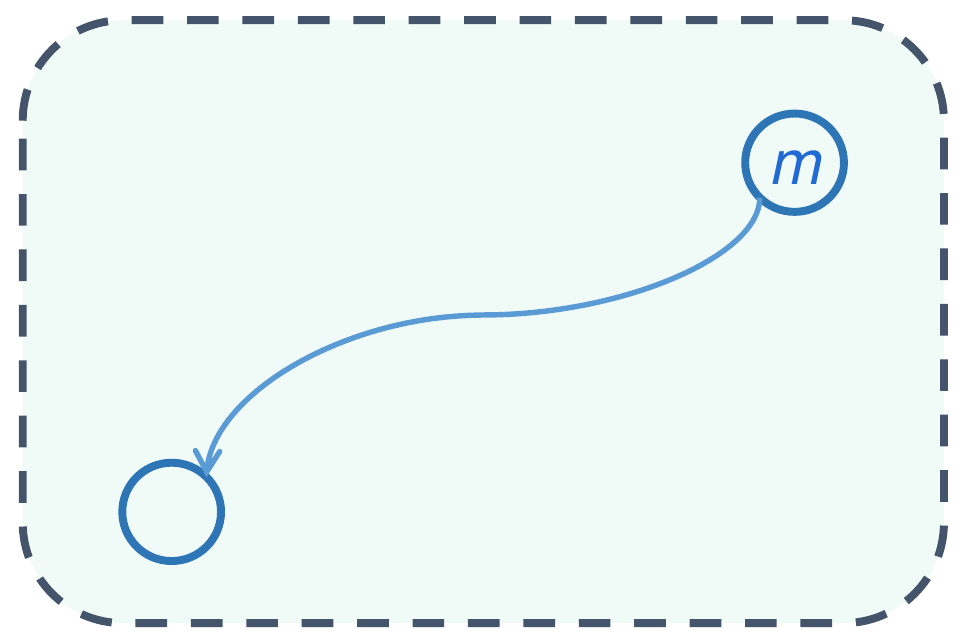}
\subcaption{}
\label{optim_1}
\end{minipage}
\begin{minipage}[ht]{0.49\linewidth}
\includegraphics[width=4.2cm]{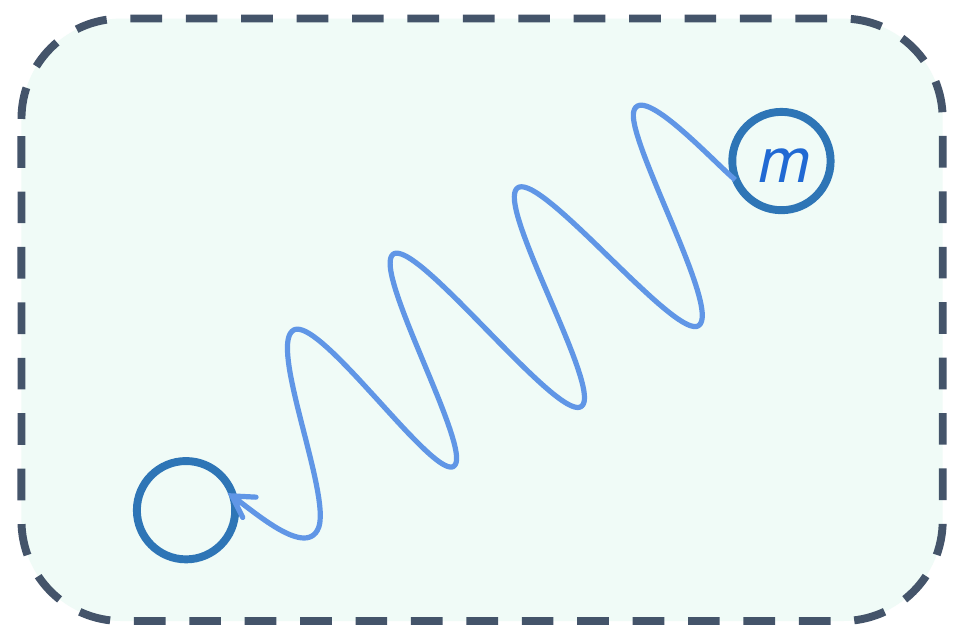}
\subcaption{}
\label{optim_2}
\end{minipage}
\begin{minipage}[ht]{0.48\linewidth}
\includegraphics[width=4.2cm]{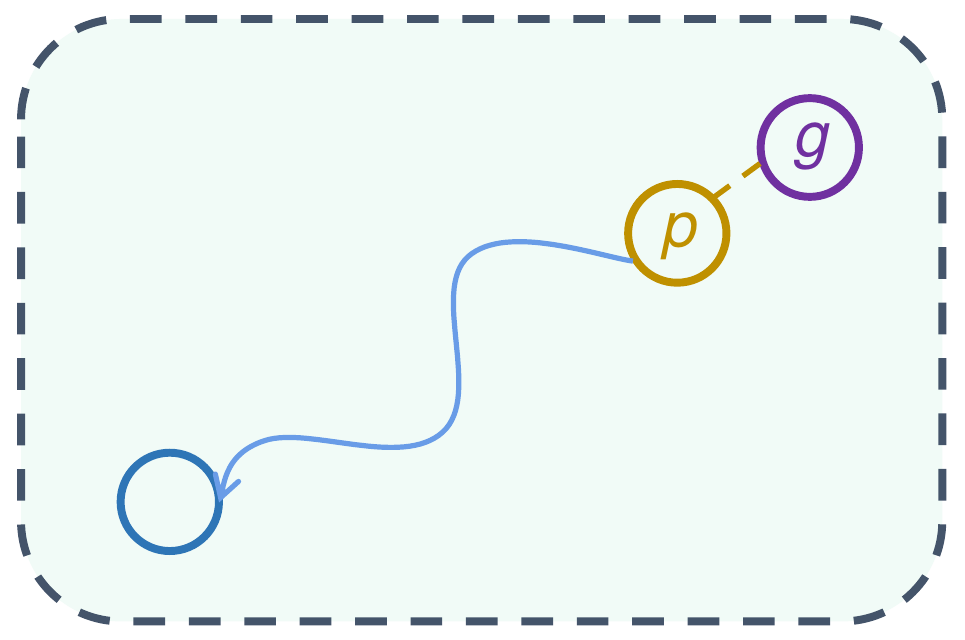}
\subcaption{}
\label{optim_3}
\end{minipage}
\begin{minipage}[ht]{0.49\linewidth}
\includegraphics[width=4.2cm]{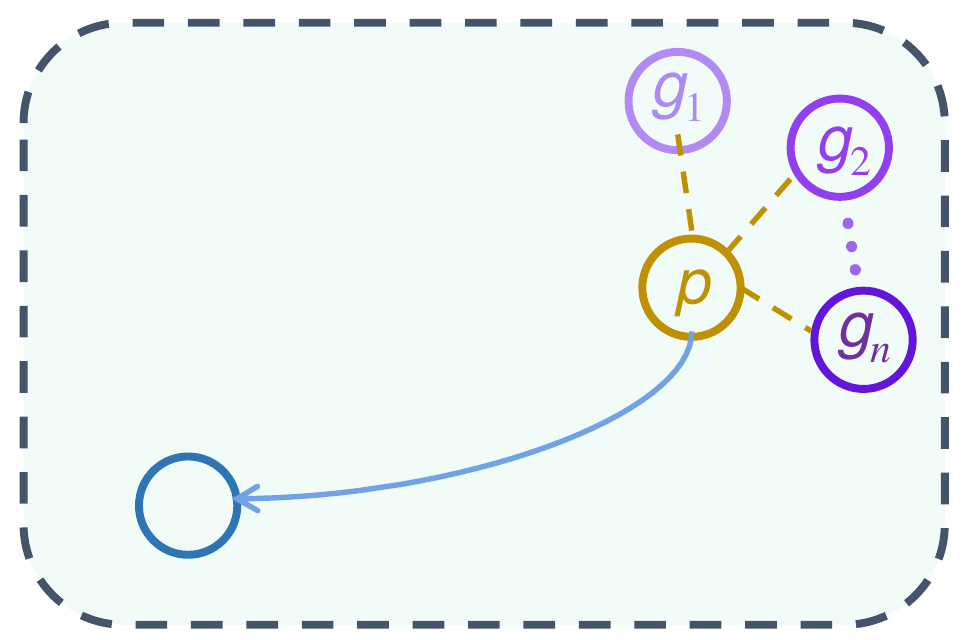}
\subcaption{}
\label{optim_4}
\end{minipage}
\end{center}

\caption{
Diagram of training landscape of conventional training, SST, and MASST. The hollow circle denotes optimal solution, ``$m$" denotes the model of conventional training, ``$p$" denotes the probe network, ``$g$" denotes the gallery network, ``$g_1$", ``$g_2$", $\cdots$, ``$g_n$" denote multiple agents of gallery network. (a) Conventional training on deep data. (b) Conventional training on long-tail data. (c) SST on long-tail data. (d) MASST on long-tail data. By replacing the single gallery network in SST with multiple agents, MASST enables a smooth and steady optimization process on long-tail data.}
\label{conv&sst&masst}
\end{figure}

According to the above experimental results, we further intuitively summarize the optimization processes of conventional training, SST, and MASST in Fig.~\ref{conv&sst&masst}.
The optimization process of conventional training on deep data is smooth.
However, when encountering long-tail data, the process to obtain an optimal solution becomes tortuous.
This issue has been greatly improved in SST.
Fortunately, with the combination of the gallery network stack, MASST achieves the smoothest optimization process.

\begin{table}[ht]
	\setlength{\tabcolsep}{0.3cm}
	\begin{center}
	\caption{
	Average (AVG.) and standard deviation (SD.) of curvature of training landscape by the conventional training, SST, and MASST along with the training iteration. 
	One can refer to Fig.~\ref{sn} for the trend of curvature along with the training iteration.
	}
	\label{tb:sn}
	\begin{tabular}{|c|c|c|c|c|}
	\hline
    \multirow{2}{*}{Method}&
    \multicolumn{2}{c|}{Shallow Face Data}&
    \multicolumn{2}{c|}{Long-tail Face Data}
    \\ \cline{2-5}& AVG. & SD. & AVG. & SD.\\
    \hline\hline
    Convt.&5.7518&1.5522&19.9664&16.5690\\
    SST&11.6319&0.4789&12.0648&1.6331\\
    MASST&\textbf{9.8915}&\textbf{0.4484}&\textbf{9.4698}&\textbf{1.0015}\\
    \hline
    \end{tabular}
	\end{center}
\end{table}

\section{Experiments}
\label{EXPERIMENTS}
This section is structured as follows.
Section~\ref{section_exp_datasets} introduces the datasets and experimental settings.
Section~\ref{section_ablation_study} includes the ablation study on MASST.
Section~\ref{section_masst_on_shallow_data} demonstrates the significant improvement by MASST on shallow face data with various loss functions (Section~\ref{section_masst_with_various_losses}) and various backbones (Section~\ref{section_masst_with_various_nets}).
Section~\ref{section_masst_on_long_tail_data} verifies MASST performs well on long-tail face data.
Section~\ref{section_masst_on_deep_data} shows MASST can achieve leading performance on deep face data.
Section~\ref{section_masst_on_domain_transfer} studies MASST also outperforms conventional training for the domain transfer.

\subsection{Datasets and Experimental Settings}
\label{section_exp_datasets}

{\bf \noindent Training Data} 
We use the public datasets to prove the reproducibility.
To obtain shallow data, two images are randomly selected for each ID from the MS1M-v1c~\cite{trillionpairs.org} dataset. 
Thus, the shallow data includes 72,778 IDs and 145,556 images.
And then, we sample from the MS1M-v1c dataset to construct long-tail data.
The sampling method is concretely expressed in Section~\ref{MASST for long-tail data}.
The samples in head classes are sufficient.
Closer to the tail classes, the number of samples per ID goes down but there are at least two images as input for training.
For deep data, we employ the full MS1M-v1c which has 44 images per ID on average.
As a real-world surveillance face recognition benchmark, QMUL-SurvFace~\cite{cheng2018surveillance} is utilized for the experiment of pretrain-finetune as well.

{\bf \noindent Test Data} 
To evaluate comprehensively, we adopt LFW~\cite{huang2008labeled}, BLUFR~\cite{liao2014benchmark}, AgeDB-30~\cite{moschoglou2017agedb}, CFP-FP~\cite{sengupta2016frontal}, CALFW~\cite{zheng2017cross}, CPLFW~\cite{zheng2018cross}, MegaFace~\cite{kemelmacher2016megaface},
IJB-B~\cite{whitelam2017iarpa}, IJB-C~\cite{maze2018iarpa}
and QMUL-SurvFace~\cite{cheng2018surveillance} datasets. 
AgeDB-30 and CALFW concentrate on cross-age face verification. 
CFP-FP and CPLFW aim at face verification with different poses. 
Based on the LFW dataset, BLUFR is a benchmark protocol that contains verification (VR) and open-set identification (DIR) scenarios, with a focus on the low false accept rate (FAR). 
VR @FAR=1e-5 is adopted for performance measurement in our experiments.
MegaFace evaluates the performance of face recognition at the million scales of distractors. 
IJB-C is a further extension of IJB-B (the extension of IJB-A~\cite{klare2015pushing}), the former having more subjects with still images and more frames from videos.
Compared with the above benchmark, the QMUL-SurvFace test set places emphasis on real-world surveillance face recognition.

{\bf \noindent Prepossessing} 
Detection is carried out on each original image by the FaceBoxes~\cite{zhang2017faceboxes}. 
Then, we employ five facial landmarks~\cite{feng2018wing} for alignment and cropping to 144$\times$144 RGB images and augment data through resizing, rotation, grayscale conversion, and horizontal flipping. 

{\bf \noindent CNN Architecture} 
To reduce time overhead while ensuring good performance, MobileFaceNet~\cite{chen2018mobilefacenets} is chosen in the ablation study and the experiments with various loss functions. 
Besides, we use Attention-56~\cite{wang2017residual} in the synthetic long-tail data, deep data, and transfer learning. 
The output is a 512-dimension feature. 
In addition, we adopt extra backbones to prove the convergence of MASST with various architectures, including VGG-16~\cite{Simonyan2014Very}, SE-ResNet-18~\cite{hu2018squeeze}, ResNet-50 and -101~\cite{he2016deep}.

{\bf \noindent Training and Evaluation} 
We employ four NVIDIA Tesla P40 GPUs for training.
The batch size is 256 and the learning rate begins with 0.05. 
For the shallow data and long-tail data, the learning rate is divided by 10 at the 36k, 54k iterations and the training process is finished at 64k iterations.
For the deep data, the learning rate is divided at the 72k, 96k, 112k iterations, and finished at 120k iterations. 
For transfer learning, the learning rate starts from 0.001, and we divide it by 10 at the 6k, 9k iterations, and finish at 10k iterations.
The momentum is 0.9 and the weight decay is 5e-4.
The number of classes in training datasets determines the size of the gallery queue, so we empirically set it as 16,384 for shallow, long-tail, and deep data, and 2,560 for QMUL-SurvFace. 
The scale parameter $s$ is fixed as 30 and the number of gallery agents $n$ in MASST is set to 3.
In the evaluation stage, the last layer output from the probe network is extracted as the face representation. 
In addition, we utilize the cosine similarity as the similarity metric. 
In order to evaluate strictly and precisely, we remove all the overlapping IDs between training and test datasets according to the list~\cite{wang2019co}.

{\bf \noindent Loss Function} 
The proposed method can be flexible integrated with various training loss functions. 
We choose classification and embedding learning loss functions as the baseline simultaneously, and then integrate them with MASST to make comparisons.
The classification loss functions include A-softmax~\cite{liu2017sphereface}, AM-softmax~\cite{wang2018additive}, Arc-softmax~\cite{deng2019arcface}, AdaCos~\cite{zhang2019adacos}, MV-softmax~\cite{wang2020mis},  DP-softmax~\cite{zhu2019large} and Center loss~\cite{wen2016discriminative}. 
The embedding learning methods include Contrastive~\cite{sun2014deep}, Triplet~\cite{schroff2015facenet} and N-pairs~\cite{sohn2016improved}. 

\begin{table}[ht]
	\setlength{\tabcolsep}{0.13cm}
	\begin{center}
		\caption{
		        Ablation study on shallow data. Performance ($\%$) on LFW, AgeDB-30, CFP-FP, CALFW, CPLFW and BLUFR (VR @FAR=1e-5).		}
		\label{tbl:ablation_study}
		\begin{tabular}{|c||c|c|c|c|c|c|}
			\hline
     &~LFW~& AgeDB &~CFP~ & CALFW & CPLFW & BLUFR\\
    \hline\hline
    \multicolumn{7}{|c|}{softmax}\\
    \hline
         Convt.& 92.64 & 73.96 & 70.80 & 73.05 & 62.64 & 27.05\\
         \hline
         A & 91.36 & 71.85 & 69.00 & 72.14 & 61.35 & 24.87\\
         \hline
         B & 93.43 & 76.00 & 71.46 & 74.65 & 62.68 & 30.65\\
         \hline
         C & 96.62 & 82.63 & 79.10 & 80.18 & 67.55 & 52.05\\
         \hline
         D & 98.32 & 88.77 & 84.81 & 86.63 & 74.80 & 69.93\\
         \hline
         SST &98.77 & 91.60 & \textcolor{blue}{\textbf{88.63}} & \textcolor{blue}{\textbf{89.82}} & 78.43 & 77.58\\
         \hline
         MASST-MA &98.80&91.15&87.70&89.62&77.92&78.42 \\
        \hline
         MASST &\textcolor{blue}{\textbf{98.85}}&\textcolor{blue}{\textbf{91.63}}&88.13&89.80&\textcolor{blue}{\textbf{78.53}} &\textcolor{blue}{\textbf{81.53}} \\
    \hline\hline
    \multicolumn{7}{|c|}{A-softmax}\\
    \hline
         Convt.&94.67 &77.88 & 72.90 & 75.85 & 64.00 & 37.16\\
         \hline
         A & 93.76 & 76.79 & 71.35 & 74.56 & 62.80 & 35.18\\
         \hline
         B & 94.62 & 78.08 & 74.03 & 76.35 & 63.87 & 38.35\\
         \hline
         C & 96.32 & 82.28 & 81.30 & 81.05 & 68.77 & 57.13\\
         \hline
         D & 97.52 & 85.83 & 81.87 & 83.88 & 71.03 & 60.79\\
         \hline
         SST & 98.98 &\textcolor{blue}{\textbf{91.88}} & \textcolor{blue}{\textbf{89.54}} & 89.73 & 77.68 & 80.65\\
         \hline
         MASST-MA &98.87&91.05&88.47&89.47&77.92&79.71 \\
         \hline 
         MASST & \textcolor{blue}{\textbf{99.03}} & 91.72 &89.29 &\textcolor{blue}{\textbf{90.03}} & \textcolor{blue}{\textbf{78.02}} &\textcolor{blue}{\textbf{82.45}} \\
    \hline\hline
    \multicolumn{7}{|c|}{AM-softmax}\\
    \hline
        Convt.& 92.75 & 75.30 & 68.74 & 76.63 & 63.63 & 33.23\\
        \hline
        A & 92.35 & 74.12 & 68.08 & 74.89 & 62.76 & 32.12\\
         \hline
        B & 93.25 & 76.16 & 69.17 & 77.78 & 63.88 & 36.59\\
         \hline
        C & 98.02 & 86.37 & 85.17 & 85.72 & 72.83 & 62.07\\
         \hline
        D & 98.30 & 88.18 & 87.31 & 87.93 & 76.27 & 75.46\\
        \hline
        SST & 98.97 & 92.25 &88.97 & 90.23 & 79.45& 84.95\\
        \hline
         MASST-MA &98.97&91.63&87.89&89.88&79.32&86.74 \\
        \hline 
         MASST &\textcolor{blue}{\textbf{99.12}} &\textcolor{blue}{\textbf{92.73}} &\textcolor{blue}{\textbf{89.23}}  &\textcolor{blue}{\textbf{90.58}} &\textcolor{blue}{\textbf{79.82}} & \textcolor{blue}{\textbf{86.79}}\\
     \hline\hline
    \multicolumn{7}{|c|}{Arc-softmax}\\
    \hline
        Convt.& 94.32 & 77.80 & 71.25 & 78.15 & 65.45 & 40.34\\
        \hline
        A & 93.60 & 77.35 & 70.59 & 77.78 & 64.28& 40.08\\
         \hline
        B & 94.48 & 78.42 & 72.15 & 78.65 & 65.78 & 42.50\\
         \hline
        C & 98.20 & 85.28 & 81.50 & 83.50 & 71.32 & 60.67\\
        \hline
        D & 98.08 & 88.68 & 84.54 & 86.92 & 74.40 & 68.84\\
        \hline
        SST & 98.95 & 91.73 & 88.59 & 89.85 &\textcolor{blue}{\textbf{79.60}} & 82.74\\
        \hline
         MASST-MA &98.97&91.18&88.70&89.53&78.88&82.68 \\
        \hline
         MASST & \textcolor{blue}{\textbf{99.10}} &\textcolor{blue}{\textbf{92.08}} & \textcolor{blue}{\textbf{89.24}} & \textcolor{blue}{\textbf{90.52}} & 79.40 &\textcolor{blue}{\textbf{84.67}} \\
         \hline
		\end{tabular}
	\end{center}
\end{table}

\begin{figure}[ht]
    \centering
    \includegraphics[scale=0.45]{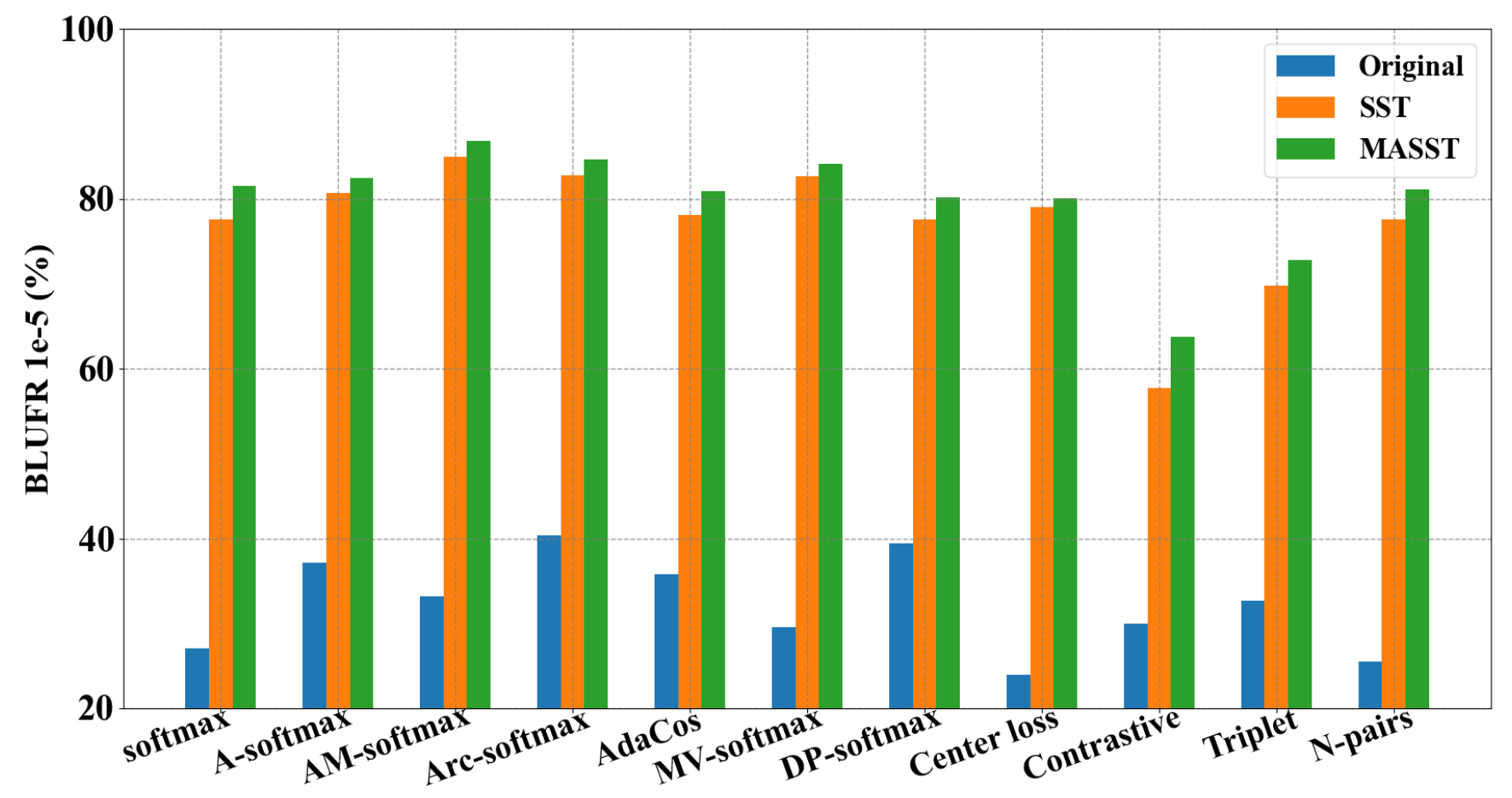}
    \caption{After being integrated with MASST, every loss function obtains a large increase in shallow face learning.
The blue bars correspond to the conventional training on shallow data, the orange bars correspond to SST on shallow data, the green bars correspond to MASST on shallow data. 
The test results are the verification rates at FAR=1e-5 on BLUFR.
Best viewed in color.}
    \label{various_loss}
\end{figure}

\begin{figure}[ht]
    \centering
    \includegraphics[scale=0.36]{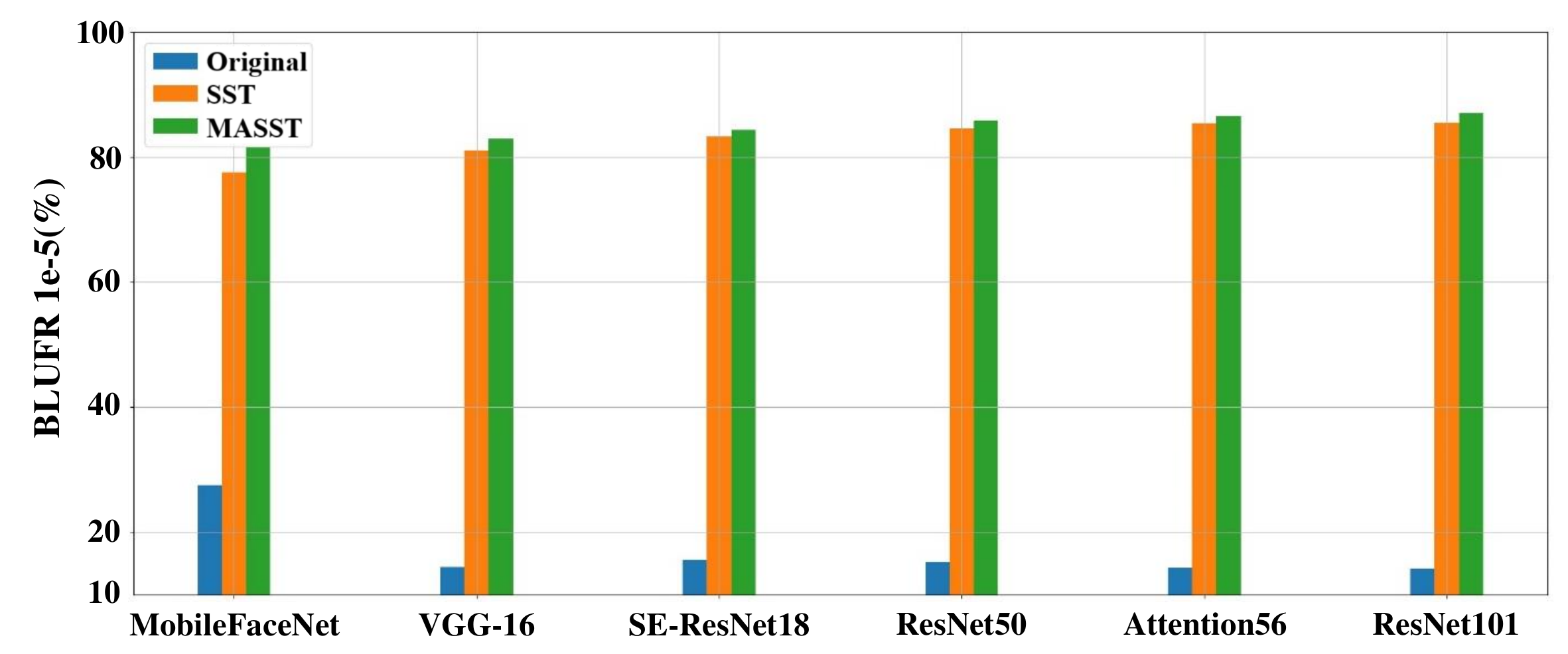}
    \caption{The test accuracy of each network on BLUFR. The blue bars refer to the network trained by conventional training, the orange bars refer to the network trained by SST, the green bars refer to the network trained by MASST. Best viewed in color.}
    \label{performance_scatter}
\end{figure}

\subsection{Ablation Study}
\label{section_ablation_study}
To analyze comprehensively, we compare SST and MASST with the different choices aforementioned, such as the network constraint ($\| \phi_g - \phi_p \| <  \epsilon'$) and the prototype constraint ($\beta (\alpha - \|w_y\|)$).
Table~\ref{tbl:ablation_study} compares their performance with four basic loss functions (softmax, A-Softmax, AM-Softmax and Arc-softmax).
In this table, ``Convt." denotes the plain training, 
``A" denotes the prototype constraint, 
``B" denotes the network constraint, 
``C" denotes the gallery queue, 
``D" denotes the combination of ``B" and ``C", 
``SST" denotes the ultimate scheme of Semi-Siamese Training which includes the moving-average updating Semi-Siamese networks and the training scheme with gallery queue,
``MASST-MA" denotes the ultimate scheme of Multi-Agent Semi-Siamese Training which replaces the single gallery network in SST with a gallery network stack that involves multiple agents and updates by the moving-average.
``MASST" denotes the gallery network stack in MASST updates in a composite way.
From Table~\ref{tbl:ablation_study}, we can conclude: (1) enlarging $w_y$ in every dimension indiscriminately is not effective in shallow face learning, as the prototype constraint ``A" leads to decrease in most terms; (2) compared with the network constraint ``B" and the gallery queue ``C", the combination of them ``D" obtains further improvement; (3) SST employs moving-average updating and gallery queue, and gets better performance; (4) finally, MASST sets up the gallery network stack including multiple agents and designs a more appropriate update mode, which achieves the best results by most of the terms.
The comparison indicates MASST can address the problem in shallow face learning and obtain a slight improvement in most test accuracy compared with SST.

\begin{table}[ht]
	\setlength{\tabcolsep}{0.04cm}
	\begin{center}
	\caption{
	Comparison of MASST and SST on synthetic long-tail data training with MobileFaceNet ($r \in (0,0.1]$). In MegaFace, ``Id." refers to face identification rank1 accuracy with 1M distractors, and ``Veri." refers to face verification rate at 1e-6 FAR.
	}
	\label{tb:results_on_long-tail_data_0.1}
	\begin{tabular}{|c|c|c|c|c|c|c|c|c|c|}
	\hline
	\multirow{2}*{$r$}& 
	\multirow{2}*{Method}& 
    \multirow{2}*{LFW}&
    \multirow{2}*{AgeDB}&
    \multirow{2}*{CFP}&
    \multirow{2}*{CALFW}&
    \multirow{2}*{CPLFW}&
    \multirow{2}*{BLUFR}&
    \multicolumn{2}{c|}{MegaFace}
    \\ \cline{9-10}&&&&&&&& Id. &Veri.\\
    \hline\hline
\multirow{2}*{0.02}&SST&99.38&93.37&92.34&91.87&\textcolor{blue}{\textbf{82.58}}&87.10&77.78&82.77\\
{}&MASST&\textcolor{blue}{\textbf{99.50}}&\textcolor{blue}{\textbf{94.70}}&\textcolor{blue}{\textbf{93.10}}&\textcolor{blue}{\textbf{92.45}}&82.45&\textcolor{blue}{\textbf{91.58}}&\textcolor{blue}{\textbf{80.61}}&\textcolor{blue}{\textbf{83.76}}\\
    \hline\hline
\multirow{2}*{0.04}&SST&99.33&\textcolor{blue}{\textbf{93.92}}&92.53&91.78&\textcolor{blue}{\textbf{82.70}}&88.77&78.74&82.73\\
{}&MASST&\textcolor{blue}{\textbf{99.50}}&93.85&\textcolor{blue}{\textbf{92.86}}&\textcolor{blue}{\textbf{92.58}}&82.48&\textcolor{blue}{\textbf{89.22}}&\textcolor{blue}{\textbf{79.92}}&\textcolor{blue}{\textbf{83.31}}\\
    \hline\hline
\multirow{2}*{0.06}&SST&99.35&93.83&91.90&\textcolor{blue}{\textbf{92.20}}&82.00&87.07&77.78&81.92\\
{}&MASST&\textcolor{blue}{\textbf{99.40}}&\textcolor{blue}{\textbf{94.25}}&\textcolor{blue}{\textbf{92.86}}&91.87&\textcolor{blue}{\textbf{82.13}}&\textcolor{blue}{\textbf{87.48}}&\textcolor{blue}{\textbf{79.97}}&\textcolor{blue}{\textbf{82.59}}\\
    \hline\hline
\multirow{2}*{0.08}&SST&99.37&93.60&91.81&\textcolor{blue}{\textbf{92.18}}&82.08&87.65&76.39&82.60\\
{}&MASST&\textcolor{blue}{\textbf{99.55}}&\textcolor{blue}{\textbf{93.85}}&\textcolor{blue}{\textbf{92.56}}&91.95&\textcolor{blue}{\textbf{82.23}}&\textcolor{blue}{\textbf{89.52}}&\textcolor{blue}{\textbf{79.39}}&\textcolor{blue}{\textbf{82.73}}\\
    \hline\hline
\multirow{2}*{0.1}&SST&99.38&93.85&\textcolor{blue}{\textbf{92.09}}&91.80&\textcolor{blue}{\textbf{81.97}}&\textcolor{blue}{\textbf{89.05}}&76.48&80.20\\
{}&MASST&\textcolor{blue}{\textbf{99.43}}&\textcolor{blue}{\textbf{94.02}}&91.81&\textcolor{blue}{\textbf{92.33}}&81.78&88.72&\textcolor{blue}{\textbf{77.90}}&\textcolor{blue}{\textbf{83.47}}\\
\hline 
\end{tabular}
\end{center}
\end{table}

\begin{table}[ht]
	\setlength{\tabcolsep}{0.2cm}
	\begin{center}
		\caption{
		        Performance ($\%$) on MegaFace (face verification rate at 1e-6 FAR) as the degree of long-tail distribution intensifies.		
		        }
		\label{tbl:alpha0-0.3}
		\begin{tabular}{|c||c|c|c|c|c|c|c|}
			\hline
     $r$&~0~& 0.05 &~0.10~ & 0.15 & 0.20 & 0.25 & 0.30\\
    \hline\hline
    \multicolumn{8}{|c|}{softmax}\\
    \hline
        Convt.& 92.00 & 90.70 & 86.68 & 82.43 & 64.10 & 41.20 & 21.49\\
        \hline
        SST &93.23 & 91.37 & 89.62 & 86.62 & 86.03 & 85.68 & \textcolor{blue}{\textbf{80.90}}\\
        \hline
        MASST &\textcolor{blue}{\textbf{94.76}}&\textcolor{blue}{\textbf{92.95}}& \textcolor{blue}{\textbf{90.57}} & \textcolor{blue}{\textbf{87.13}} &\textcolor{blue}{\textbf{86.89}} &\textcolor{blue}{\textbf{86.57}} & 80.05\\
    \hline\hline
    \multicolumn{8}{|c|}{AM-softmax}\\
    \hline
        Convt.& 96.35 & 94.82 & 93.67 & 91.60 & 89.57 & 69.42 & 39.68\\
        \hline
        SST & 96.96 & 95.67 & 93.96 & \textcolor{blue}{\textbf{91.94}} & 89.76 & 87.14 & 85.61\\
        \hline 
        MASST &\textcolor{blue}{\textbf{97.18}} &\textcolor{blue}{\textbf{96.40}} &\textcolor{blue}{\textbf{94.42}} & 91.83 &\textcolor{blue}{\textbf{90.18}} & \textcolor{blue}{\textbf{87.68}} & \textcolor{blue}{\textbf{85.90}} \\
     \hline\hline
    \multicolumn{8}{|c|}{Arc-softmax}\\
    \hline
        Convt.& \textcolor{blue}{\textbf{96.81}} & \textcolor{blue}{\textbf{95.73}} & 93.82 & 91.73 & 87.65 & 77.55 & 38.82\\
        \hline
        SST & 96.50 & 95.71 & \textcolor{blue}{\textbf{94.95}} & 92.71 & \textcolor{blue}{\textbf{88.02}} & 86.94 & 80.46\\
        \hline
        MASST & 96.67 & 95.56 & 94.57 &\textcolor{blue}{\textbf{93.16}} & 88.01 & \textcolor{blue}{\textbf{87.46}} & \textcolor{blue}{\textbf{82.12}} \\
        \hline
		\end{tabular}
	\end{center}
\end{table}

\subsection{Advantage of MASST on Shallow Data}
\label{section_masst_on_shallow_data}
To prove the effectiveness of MASST on shallow data learning, we not only train the network with various loss functions, but also employ it to train different CNN architectures.
\subsubsection{MASST with Various Loss Functions}
\label{section_masst_with_various_losses}

First, we train the network on the shallow data with various loss functions and test it on BLUFR at FAR=1e-5 (the blue bars in Fig.~\ref{various_loss}).
The loss functions include classification and embedding ones such as softmax, A-softmax, AM-softmax, Arc-softmax, AdaCos, MV-softmax, DP-softmax, Center loss, Contrastive, Triplet, and N-pairs. 
Then, we train the same network with the same loss functions on the shallow data, but with SST scheme and MASST scheme respectively. 
As shown in Fig.~\ref{various_loss}, MASST can be flexibly integrated with every loss function.
Compared with SST (the orange bars), MASST obtains a larger increase for shallow face learning (the green bars).
Moreover, MASST is proved to be effective when facing the hard example mining strategies by the result of training on MV-softmax and embedding losses.

\subsubsection{MASST with Various Network Architectures}
\label{section_masst_with_various_nets}


\begin{table*}[htb]
	\setlength{\tabcolsep}{0.24cm}
	\begin{center}
	\caption{
	Comparison of MASST, SST, and the conventional training on deep data with various loss settings. 
	}
	\label{tb2:results_on_deep_data}
	\begin{tabular}{|c|p{0.9cm}<{\centering}|p{1.1cm}<{\centering}|p{0.9cm}<{\centering}|p{1.2cm}<{\centering}|p{1.2cm}<{\centering}|p{1.2cm}<{\centering}|p{1cm}<{\centering}|}
	\hline
    \multirow{2}*{Method} & 
    \multirow{2}*{LFW}&
    \multirow{2}*{AgeDB}&
    \multirow{2}*{CFP}&
    \multirow{2}*{CALFW}&
    \multirow{2}*{CPLFW}&
    \multicolumn{2}{c|}{MegaFace}
    \\ \cline{7-8}&&&&&& Id. &Veri.\\
    \hline\hline
Convt.(softmax)&99.58&95.33&92.66&93.18&84.47&89.89&92.00\\
Convt.(AM-softmax)~\cite{wang2018additive}&99.70&97.03&94.17&94.41&87.00&95.67&96.35\\
Convt.(Arc-softmax)~\cite{deng2019arcface}&99.73&97.18&94.37&\textcolor{blue}{\textbf{95.25}}& 87.05&96.10&96.81\\
Convt.(DP-softmax)~\cite{zhu2019large}&99.63&95.68&91.74&93.03&83.88&89.27&90.94\\
Convt.(Contrastive)~\cite{sun2014deep}&99.50&92.91&91.76&87.56&80.13&60.14&63.59\\
Convt.(Triplet)~\cite{schroff2015facenet}&99.47&93.32&94.49&89.32&82.25&65.65&69.18\\
Convt.(N-pairs)~\cite{sohn2016improved}&99.53&94.58&93.43&92.10&83.15&76.87&78.28\\
\hline\hline
SST (softmax)&99.67&96.37&94.96&94.18&85.82&91.01&93.23\\
SST (AM-softmax)&99.75&97.20&95.10&94.62&88.35& \textcolor{blue}{\textbf{96.27}}&96.96\\
SST (Arc-softmax)& \textcolor{blue}{\textbf{99.77}}&97.12&95.96&94.78&87.15&95.63&96.50\\
SST (DP-softmax)&99.68&96.24&94.56&93.78&86.04&92.08&93.57\\
SST (Contrastive)&99.56&93.14&92.71&92.13&81.78&77.59&82.44\\
SST (Triplet)&99.50&94.30&93.30&92.05&82.67&81.76&83.29\\
SST (N-pairs)&99.65&96.12&94.86&94.32&84.74&91.72&93.48\\
\hline\hline 
MASST (softmax)&99.67&97.23&95.37&94.97&87.05&92.64&94.76\\
MASST (AM-softmax)&\textcolor{blue}{\textbf{99.77}}&\textcolor{blue}{\textbf{97.70}}&95.61&95.10&\textcolor{blue}{\textbf{88.63}}&96.15&\textcolor{blue}{\textbf{97.18}}\\
MASST (Arc-softmax)&99.76&97.37&\textcolor{blue}{\textbf{96.23}}&95.08&87.95&95.86&96.67\\
MASST (DP-softmax)&99.73&97.26&95.17&94.93&87.39&93.70&94.73\\
MASST (Contrastive)&99.57&93.88&93.21&91.70&83.27&81.63&84.52\\
MASST (Triplet)&99.58&94.62&94.19&92.32&84.54&84.43&86.31\\
MASST (N-pairs)&99.65&96.73 &95.24&95.02&86.18&93.36&94.89\\
\hline 
\end{tabular}
	\end{center}
\end{table*}

\begin{table*}[t!]
\begin{center}
\setlength{\tabcolsep}{0.2cm}
\caption{QMUL-SurvFace evaluation. ``TPR(\%)@FAR" includes the true positive verification rate at varying FARs, and ``TPIR20(\%)@FPIR" includes rank-20 true positive identification rate at varying false positive identification rates. 
}
\label{Finetune}
\begin{tabular}{|c|p{1cm}<{\centering}|p{1cm}<{\centering}|p{1cm}<{\centering}|p{1cm}<{\centering}|p{1cm}<{\centering}|p{1cm}<{\centering}|p{1cm}<{\centering}|p{1cm}<{\centering}|}
\hline
\multirow{2}{*}{Method}&
\multicolumn{4}{c|}{TPR(\%)@FAR}&
\multicolumn{4}{c|}{TPIR20(\%)@FPIR}
\\ \cline{2-9}&0.3&0.1&0.01&0.001&0.3&0.2&0.1&0.01\\
\hline\hline
Convt.(softmax) &73.09&52.29&26.07&12.54&8.09&6.25&3.98&1.13\\
Convt.(AM-softmax)~\cite{wang2018additive} &69.59&47.67&23.90&13.24&9.07&7.14&4.65&1.34\\
Convt.(Arc-softmax)~\cite{deng2019arcface} &68.14&48.65&24.12&11.34&8.77&6.88&4.79&1.36\\
Convt.(DP-softmax)~\cite{zhu2019large} &76.32&55.85&25.32&11.64&7.50&5.38&3.38&0.95\\
Convt.(Contrastive)~\cite{sun2014deep} &84.48&67.99&31.87&5.31&9.16&6.91&4.44&0.10\\
Convt.(Triplet)~\cite{schroff2015facenet}
&85.59&69.61&33.76&7.20&10.14&7.70&4.75&0.37\\
Convt.(N-pairs)~\cite{sohn2016improved}
&87.26&67.04&29.67&12.07&10.75&8.09&4.87&0.41\\
\hline
SST(softmax) &81.08&63.41&34.20&19.03&11.24&8.49&5.28&1.21\\
SST(AM-softmax)  &86.49&69.41&36.21&18.51&12.22&9.51&5.85&1.63\\
SST(Arc-softmax) &87.00&68.21&35.72&\textcolor{blue}{\textbf{22.18}}&12.38&9.71&6.61&\textcolor{blue}{\textbf{1.72}}\\
SST(DP-softmax) &87.69&69.69&36.32&14.83&10.20&7.83&5.14&1.08\\
SST(Contrastive) &87.54&69.91&32.15&9.58&9.87&7.38&4.76&0.78\\
SST(Triplet)  &\textcolor{blue}{\textbf{90.65}}&73.35&33.85&12.48&11.09&8.14&5.27&0.92\\
SST(N-pairs) &89.31&71.26&32.34&15.96&11.30&9.13&5.68&1.22\\
\hline
MASST(softmax) &81.78&62.87 &33.01&16.26&11.71&9.07&5.78&1.35\\
MASST(AM-softmax)  &86.70&68.75&36.13&18.72&11.94&9.18&6.12&1.65\\
MASST(Arc-softmax) &85.99&69.08&\textcolor{blue}{\textbf{37.46}}&21.97&12.27&9.21&5.64&1.68\\
MASST(DP-softmax) &86.61&67.31&36.62&20.62&11.18&8.31&5.65&1.25\\
MASST(Contrastive) &89.27&73.28&37.03&19.17&10.25&7.90&4.26&0.43\\
MASST(Triplet)  &90.10&\textcolor{blue}{\textbf{74.18}}&36.91&10.00&\textcolor{blue}{\textbf{12.47}}&\textcolor{blue}{\textbf{9.96}}&\textcolor{blue}{\textbf{6.63}}&0.63\\
MASST(N-pairs) &88.67&69.89&36.86&16.87&11.84&8.97&5.57&1.30\\
\hline

\end{tabular}
\end{center}
\end{table*}

To demonstrate the stable convergence in the training, we employ MASST to train different CNN architectures, including MobileFaceNet, VGG-16, SE-ResNet-18, Attention-56, ResNet-50, and -101. 
The test accuracy of each network on BLUFR is shown in Fig.~\ref{performance_scatter}. 
The blue bars refer to the network trained by conventional training, the green bars refer to the network trained by MASST, and the orange bars refer to the network trained by SST.
For conventional training, the test accuracy decreases with the deeper network architectures, showing that the larger model size exacerbates the model degeneration and over-fitting. 
In contrast, as the network becomes heavy, the test accuracy of SST and MASST both increase, showing that they make increasing contributions with more complicated architectures and MASST shows advantages compared with SST.

\subsection{Advantage of MASST on Long-tail Data}
\label{section_masst_on_long_tail_data}

First, we set $r$ at 0.02 intervals between 0 and 0.1 to compare the performance of SST and MASST on LFW, AgeDB-30, CFP-FP, CALFW, CPLFW, BLUFR, and MegaFace.
In order to balance performance and the time costs, we choose softmax as the loss function and use the MobileFaceNet to conduct this experiment.
The result is shown in Table~\ref{tb:results_on_long-tail_data_0.1}.
We can find that MASST gains higher accuracy in most of the test sets such as LFW, AgeDB-30, CFP-FP, CALFW, and BLUFR.
In particular, it shows greater advantages than SST on MegaFace.


Then, we compare the performance of conventional training, SST and MASST on IJB-B and IJB-C.
The results of experiments are shown in the supplementary material.
With MASST, there are some improvements for different cases of loss functions both on IJB-B and IJB-C.


To observe the performance trends of MASST, SST, and conventional training when $r$ is changed, we set $r$ at 0.05 intervals between 0 and 0.3 (if $r > 0.3$, too few samples to get satisfied training results).
The number of images in the synthetic long-tail face data for different $r$ is shown in Fig.~\ref{long-tail data} (b).
We employ Attention-56 and train the network with various loss functions such as softmax, AM-softmax, and Arc-softmax.
Specifically, compare the performance of conventional training, SST, and MASST on the verification task of MegaFace.

As shown in Table~\ref{tbl:alpha0-0.3}, with the increase of $r$, \textit{i.e.} the degree of long-tail distribution is intensified, the verification rate of conventional training decreases significantly.
By contrast, MASST is more adaptable and achieves better performance than SST in long-tail face learning.

\subsection{Stable Advantage on Deep Data}
\label{section_masst_on_deep_data}
The previous experiments show MASST has well tackled the problems in shallow face learning and long-tail face learning and obtained significant improvement in test accuracy. 
To further explore the advantage of MASST for wider application, we adopt MASST scheme on the deep data (full version of MS1M-v1c), and make comparisons with the conventional training and SST. 
Table~\ref{tb2:results_on_deep_data} shows the performance on LFW, AgeDB-30, CFP-FP, CALFW, CPLFW and MegaFace.
MASST gains the leading accuracy in most of the test sets, and also the competitive results on CALFW and the face identification rank1 accuracy with 1M distractors.
MASST (softmax) achieves about two percent improvement than conventional training (softmax) on AgeDB-30, CFP-FP, CALFW, and CPLFW which include the hard cases of large face pose or large age gap.
Compared with SST, MASST shows better network property.
Notably, MASST and SST reduce a large number of FC parameters by which the classification loss is computed for the conventional training.

\subsection{Stable Advantage on Domain Transfer}
\label{section_masst_on_domain_transfer}
The public training datasets, such as MS-Celeb-1M and VGGFace2, are different from the captured face images in real-world scenarios of face recognition.
They are consist of well-posed face images collected from the internet.
However, in real-world applications, the external conditions are more complicated.
To address this issue, the typical routine is to pretrain a network on the public training datasets and fine-tune it on real-world face data.
Although MASST is proved to be superior in shallow and long-tail face learning, we still look forward to extending it to the challenge on domain transfer.
So, we conduct an extra experiment with pretraining on MS1M-v1c and finetuning on QMUL-SurvFace in this subsection. 
From Table~\ref{Finetune}, we can find that, no matter for classification learning or embedding learning, MASST boosts the performance significantly in both verification and identification, compared with the conventional training and SST.

\section{Conclusion}
\label{Conclusion}
In this paper, we focus on shallow and long-tail face learning.
We analyze the causes of these problems and verify the existing training methods are not effective enough to solve them.
The shallow and imbalanced data increases the training difficulty.
Besides, the feature space collapse leads to model degeneration and over-fitting.
Then, we propose Multi-Agent Semi-Siamese Training (MASST) to address these issues in shallow and long-tail face learning.
Specifically, MASST employs the Semi-Siamese networks with a gallery network stack that involves multiple agents to satisfy Lipschitz continuity and constructs the gallery queue with gallery features. 
We conduct extensive experiments to verify it can obtain leading performance when it combines with the existing loss functions and network architectures flexibly.
MASST has great improvement compared with conventional training.
The extra experiments further explore the advantage of MASST for wider applications, such as deep data training and domain transfer.




\ifCLASSOPTIONcaptionsoff
  \newpage
\fi

\bibliographystyle{IEEEtran}
\bibliography{reference}



 





\end{document}


\section{\textbf{Appendix}}
\subsection{Proof of the Formula~9}
For the clarity of the following deduction, we denote $f$ as a function of $\boldsymbol{x_g}$ and $\boldsymbol{x_p}$,
\begin{equation}
\begin{aligned}
f([\boldsymbol{x_g};\boldsymbol{x_p}]) = \frac{e^{s\,\boldsymbol{x_g}^T \boldsymbol{x_p}}}{e^{s\,\boldsymbol{x_g}^T \boldsymbol{x_p}}+\sum_{j=1,j\neq y}^{n} e^{s\,\boldsymbol{x_j}^T \boldsymbol{x_p}}}-1.
\end{aligned}
\end{equation}
As the training continues, $x_j^{\mathrm{T}}{x_p} \vert _{j \neq y}\rightarrow0$. 
We can have the approximation:
\begin{equation}
\begin{aligned}
f([\boldsymbol{x_g};\boldsymbol{x_p}]) &= \frac{e^{s\,\boldsymbol{x_g}^T \boldsymbol{x_p}}}{e^{s\,\boldsymbol{x_g}^T \boldsymbol{x_p}}+(n-1)}-1\\
&=\frac{e^{s\,\sum_{i=1}x_g^i x_p^i}}{e^{s\,\sum_{i=1}x_g^i x_p^i}+(n-1)}-1.
\end{aligned}
\end{equation}
Then derivative
\begin{equation}
\frac{\partial f}{\partial x_g^i}=\frac{s x_g^i e^{s \boldsymbol{x_g}^T \boldsymbol{x_p}}(n-1)}{[e^{s \boldsymbol{x_g}^T \boldsymbol{x_p}}+(n-1)]^2},
\end{equation}

\begin{equation}
\frac{\partial f}{\partial \boldsymbol{x_g}}=s\frac{e^{s \boldsymbol{x_g}^T \boldsymbol{x_p}}(n-1)}{[e^{s \boldsymbol{x_g}^T \boldsymbol{x_p}}+(n-1)]^2}\boldsymbol{x_g}\leq \frac{s}{2}\boldsymbol{x_g}.
\end{equation}
According to ${\Vert \boldsymbol{x_g}\Vert}_2=1$, the 2-norm of derivative
\begin{equation}
\begin{aligned}
{\Vert\nabla_{\boldsymbol{x_g}} f^{T}_{\boldsymbol{x_g}; \boldsymbol{x_p}}\Vert}_2\leq \frac{s}{2}.
\end{aligned}
\end{equation}
Therefore, the variation of $\eta_{x_g}$
\begin{equation}
\begin{aligned}
&\lVert \eta_{x_g}([\boldsymbol{x_g};\boldsymbol{x_p}])-\eta_{x_g}([\boldsymbol{x_g'};\boldsymbol{x_p}]) \rVert_2 \\
&= s\lVert \boldsymbol{x_g}-\boldsymbol{x_g'}\rVert_2 \lVert \nabla_{\boldsymbol{x_g}} f^{T}_{\boldsymbol{x_g}; \boldsymbol{x_p}}\rVert_2\\
&\leq \frac{s^2}{2}\lVert \boldsymbol{x_g}-\boldsymbol{x_g'}\rVert_2.
\end{aligned}
\end{equation}

\begin{figure}[ht]
\centering
\includegraphics[scale=0.58]{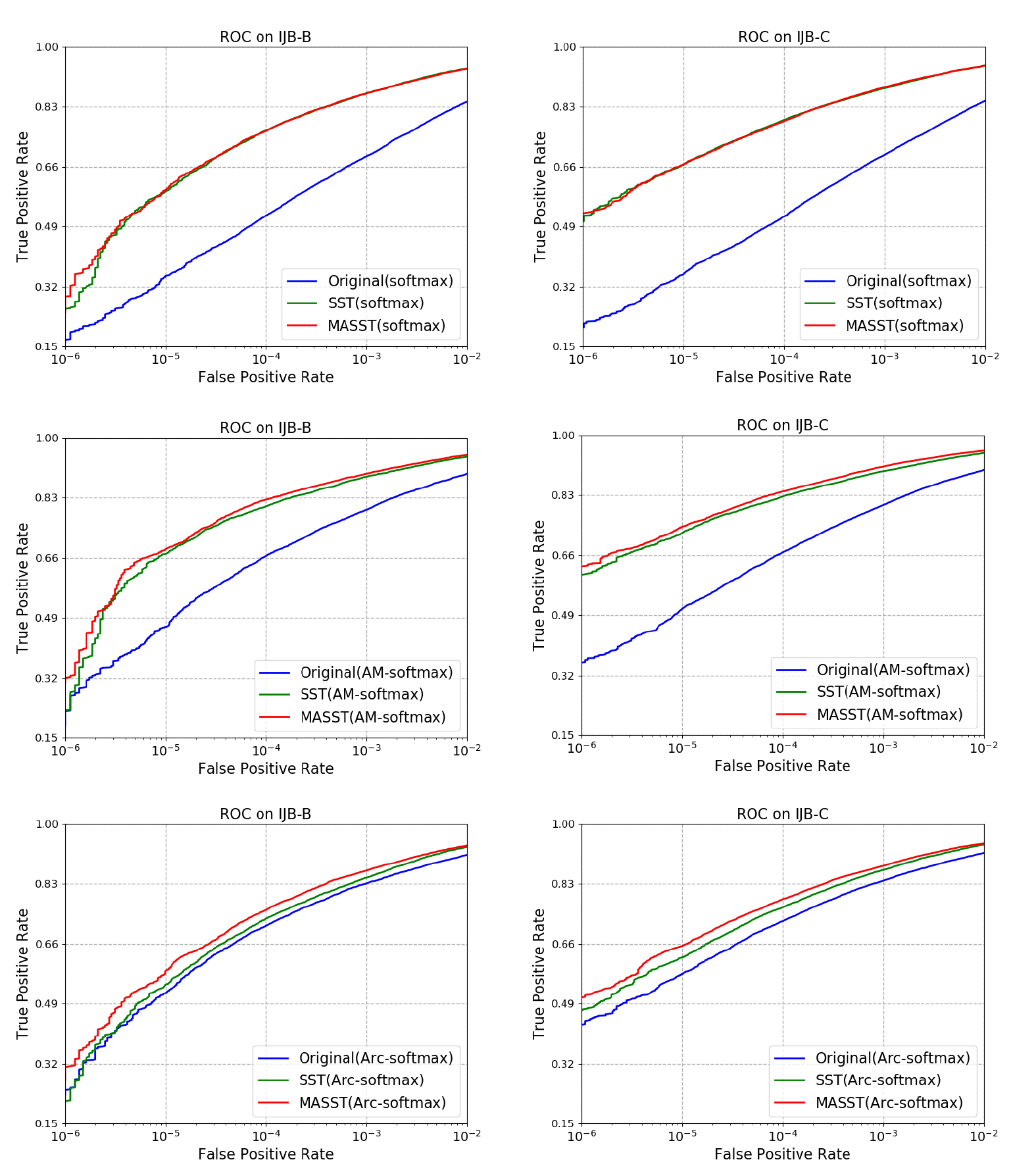}
\caption{ROC on IJB-B and IJB-C with conventional training, SST, and MASST. The top is the result of choosing softmax as the loss function. The middle is the result of choosing AM-softmax as the loss function. The bottom is the result of choosing Arc-softmax as the loss function.}
\label{IJB}
\end{figure}

\subsection{Proof of the regional monotonicity}

\begin{figure}[ht]
\centering
\includegraphics[scale=0.5]{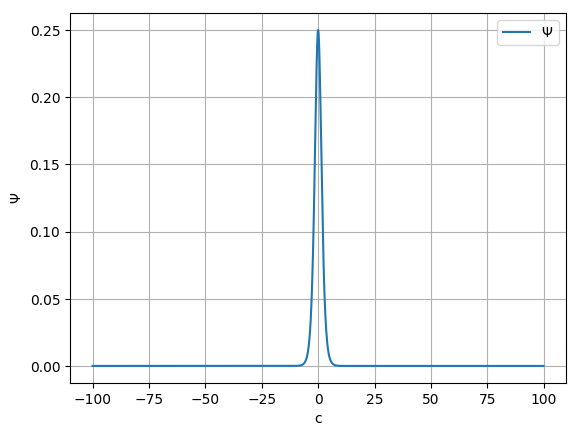}
\caption{The curve of $\Psi$.}
\label{monotonicity}
\end{figure}

According to Equation 13, we denote $\Psi$ as a function of $c$ that 
\begin{equation}
\begin{aligned}
\Psi(c)=\frac{1}{e^c+e^{-c}+2},
\end{aligned}
\end{equation}
where $c=(\boldsymbol{x_g}-\boldsymbol{x_g'})^T\boldsymbol{x_p}$. 
The curve of $\Psi$ is shown in Fig.~\ref{monotonicity}, we can find $\Psi$ is monotonically increasing when $c<0$ and monotonically decreasing when $c>0$.

\subsection{ROC on IJB-B and IJB-C}
We choose Attention-56 as the backbone and train the network on synthetic long-tail face data ($r$=0.25) with softmax, AM-softmax, and Arc-softmax respectively, to compare the performance of conventional training, SST and MASST on IJB-B and IJB-C.
The results of experiments are shown in Fig.~\ref{IJB}.
With MASST, there are some improvements for different cases of loss functions both on IJB-B and IJB-C.